\documentclass{fairmeta}
\usepackage{amsmath}
\usepackage{enumerate} 
\usepackage{bm}
\usepackage{algorithm}
\usepackage{floatflt}
\usepackage{algpseudocode}
\usepackage{amsfonts}
\usepackage{amsthm}
\usepackage{newtxtt}
\usepackage{algorithm}
\usepackage{colortbl} 
\usepackage{cleveref}
\usepackage{diagbox} 
\usepackage[utf8]{inputenc}
\usepackage{textgreek}
\usepackage{colortbl}
\usepackage{nicematrix}
\usepackage{makecell}
\usepackage{float}
\usepackage{arydshln}
\usepackage[frozencache,cachedir=.]{minted}
\usepackage{caption}
\usepackage{subcaption}
\usepackage{tcolorbox}
\usepackage{amssymb}
\usepackage{xspace}
\usepackage{wrapfig}
\usepackage{adjustbox}
\usepackage{tabularx}
\usepackage{booktabs}
\usepackage{mathtools}
\usepackage{wrapfig}
\usepackage{amssymb}
\usepackage{graphicx}
\usepackage{pifont}

\newcolumntype{C}{>{\centering\arraybackslash}X}
\newcommand{\cmark}{\ding{51}}

\usepackage{silence}
\makeatletter
\patchcmd{\wrong@fontshape}{\@gobbletwo}{}{}{}
\makeatother
\definecolor{upColor}{RGB}{17,138,21}
\definecolor{downColor}{RGB}{174,36,67}

\newtheorem{theorem}{Theorem}[]

\newtheorem{remark1}[theorem]{Remark}

\title{GeoAnchor: Collaborative Reasoning via Latent Decomposition for 3D Spatial Understanding}
\author[1,2]{Hao Li}
\author[2]{Han Fang}
\author[2,4]{Zixin Pan}
\author[2]{Xin Wei}
\author[2]{Hongbo Sun}
\author[3]{Jinglin Xu}
\author[2]{Zhiyu Lin}
\author[2]{Ye Yuan}
\author[2]{Zhongjiang He}
\author[1,*]{Yu Yu}
\author[2,*]{Hao Sun}
\affiliation[1]{Shanghai Jiao Tong University}
\affiliation[2]{Xingchen AGI Lab, China Telecom Artificial Intelligence Technology (Beijing) Co., Ltd}
\affiliation[3]{University of Science and Technology Beijing}
\affiliation[4]{The Hong Kong University of Science and Technology (Guangzhou)}
\metadata[Links]{\url{https://github.com/JerryPW/GeoAnchor}}
\date{July 14, 2026}

\begin{document}

\abstract{
Although multimodal large language models (MLLMs) have achieved remarkable progress, understanding 3D spatial relationships from 2D images remains a critical challenge. 
Existing methods primarily rely on symbolic text tokens, which inherently lack the fidelity to represent continuous geometric information.
While recent methods use latent representations to enhance reasoning, relying on a single latent type cannot adapt to the diversity of spatial tasks, leading to misalignment in complex geometric scenarios.
To address these limitations, we propose GeoAnchor, an interleaved text-latent reasoning framework. GeoAnchor decomposes 3D spatial information into three complementary components: position latents for object grounding, direction latents for relational orientation, and geometry latents for scene structure. These components are recombined in a structured space to construct local evidence while capturing global context, enabling dynamic and interpretable reasoning. Furthermore, we introduce a collaborative training strategy that guides the model from local spatial perception to comprehensive 3D understanding. Extensive experiments on diverse and complex 3D reasoning tasks demonstrate that GeoAnchor outperforms the state of the art, validating its effectiveness and generalization capabilities.
}

\maketitle

\begingroup
\renewcommand{\thefootnote}{\fnsymbol{footnote}}
\footnotetext[1]{\rmfamily\bfseries Corresponding Authors}
\endgroup

\section{Introduction}

Multimodal large language models (MLLMs) have demonstrated remarkable performance across a wide range of 2D vision-language tasks~\citep{yang2025qwen3, liu2024llavanext, deepmind2025gemini3pro_model_card}. However, a critical capability remains underdeveloped: \emph{3D spatial reasoning}.
Humans naturally infer depth, relative positions, 
and occlusions from visual scenes, yet existing MLLMs still struggle with complex spatial reasoning tasks~\citep{cheng2024spatialrgpt, zhang2025open3d}. 
This deficiency in such 3D understanding impedes their deployment in geometry-sensitive downstream applications, including robotics~\citep{kim2024openvla, pertsch2025fast}, autonomous driving~\citep{zhou2025opendrivevla, hwang2024emma, ye2025dap}, and VR/AR systems~\citep{afzal2025next}.
Existing efforts to enhance spatial reasoning capabilities generally follow two paradigms. 
The first paradigm leverages large-scale  datasets, enabling models to implicitly learn 3D relational patterns through statistical regularities in the data~\citep{cheng2024spatialrgpt, chen2024spatialvlm, li2025spatialladder, yang2025visual, cai2025scaling}. 
The second paradigm augments 2D visual inputs with explicit 3D or 2.5D signals (e.g., point clouds, depth maps, or cognitive maps), thereby providing richer global geometric structure for the model~\citep{mao2025spatiallm, zheng2025video, huang2025mllms, wu2025spatial, fan2025vlm}. 
Despite their contributions, both methods are constrained by reasoning within a discrete text space: geometric evidence is verbalized into discrete tokens, inevitably leading to the loss of fine-grained spatial cues and biasing reasoning toward linguistic priors.
Furthermore, the discrete nature of text tokens fails to capture the continuous characteristics of the physical world, rendering autoregressive generation unreliable for precise numerical tasks like distance estimation and object localization~\citep{wen2025diffusionvla}.

\begin{figure}
\includegraphics[width=\linewidth]{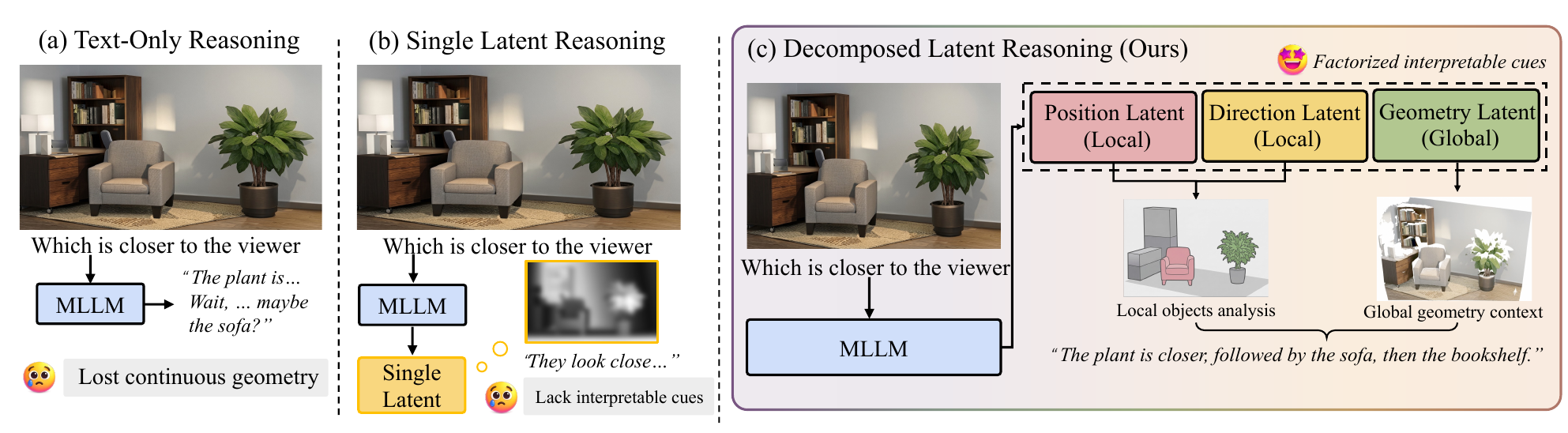}
\caption{Existing methods suffer from information loss due to verbalization or limited interpretability from entangled latents. In contrast, GeoAnchor decomposes reasoning into position, direction, and geometry factors to enable structured, interpretable local-plus-global reasoning.}
\vspace{-10pt}
\label{fig:motivation}
\end{figure}

Recent studies have explored latent reasoning to address the inherent limitations of text-based reasoning. 
By conducting intermediate reasoning within non-linguistic latent representations, rather than verbalizing every step in natural language, these methods preserve continuous semantic information, facilitating more coherent multimodal reasoning~\citep{qin2025chain, yang2025machine, li2025latent, wang2025monet}.  
When applied to spatial reasoning, researchers typically design geometrically-aware latent tokens supervised by geometric priors (e.g., depth maps)~\citep{bigverdi2025perception, liu2025ssr}.
Nevertheless, directly adopting such latent tokens for 3D spatial reasoning introduces two key limitations. 
First, existing methods rely on a single type of latent representation, which struggles to accommodate the diverse demands of spatial reasoning.
For example, while latents decoded into depth maps effectively capture near-far object relationships, they are limited in representing complex spatial information beyond depth.
Second, these latent representations primarily focus on global geometric context, offering limited insight into local object interactions. Consequently, when tasked with precise tasks such as estimating the absolute distance between two points, global latents fail to provide sufficient local evidence.

In this work, we propose \textbf{GeoAnchor}, a text–latent interleaved framework that avoids compressing geometric information into a single entangled latent representation or verbalizing it into discrete text tokens. Instead, GeoAnchor decomposes 3D spatial information into three complementary basic latent components: a \textit{position token} for precise object grounding, a \textit{direction token} for modeling relational orientation, and a \textit{geometry token} for capturing global scene structure.
The position and direction tokens provide explicit and traceable local evidence for target objects, while the geometry latent encodes global scene context. This design enables dynamic recombination of these latents within a structured space, eliminating the need for a uniform latent token design across all types of questions. Consequently, GeoAnchor achieves an interpretable latent reasoning process that adaptively accommodates the diverse demands of spatial reasoning tasks.

To enable such structured reasoning, we introduce a four-stage collaborative training strategy. 
First, we align the model with local 3D perception using large-scale object-grounding data, thereby establishing robust object-level spatial cues. 
Second, we jointly optimize local and global latent components, facilitating the evolution of reasoning from local perception to holistic spatial understanding.
Third, we refine the latent representations via text-only supervision, encouraging the model to internalize spatial reasoning logic without relying on intermediate guidance.
Finally, we incorporate reinforcement learning~\citep{shao2024deepseekmath} with pattern-specific rewards to enable adaptive selection of latent tokens, allowing the model to dynamically switch between local-only and local-plus-global reasoning patterns. 
Built on Qwen3-VL-2B~\citep{yang2025qwen3}, GeoAnchor achieves state-of-the-art performance: 68.4\% and 69.7\% accuracy on the SPAR-Bench~\citep{zhang2025flatland} and SPBench~\citep{li2025spatialladder}, respectively. 
It surpasses the base model by a significant 21.2\% margin, and outperforms GPT-4o~\citep{hurst2024gpt} and Gemini-2.5-Flash~\citep{comanici2025gemini} by 18.0\% and 13.6\%, respectively. 
Moreover, a 10.7\% performance gain on the out-of-domain ViewSpatial Bench~\citep{li2025viewspatial} confirms GeoAnchor’s robust generalization capability.

The main contributions of this paper are summarized as follows:
\begin{itemize}
    \item We propose GeoAnchor, a new 3D spatial reasoning framework that decomposes geometric reasoning into three complementary and physically meaningful latent components: position, direction, and geometry. Such a design enables dynamic latent recombination toward interpretable and query-adaptive spatial understanding.
    \item We introduce a collaborative four-stage training strategy that enables the model to perform structured local-plus-global spatial reasoning and adaptively select latent tokens.
    \item Extensive experiments on challenging 3D reasoning benchmarks, including SPAR-Bench, SPBench and ViewSpatial, demonstrate that GeoAnchor surpasses the base model by 21.2\% and outperforms GPT-4o and Gemini-2.5-Flash, showing competitive performance in 3D reasoning tasks. 
\end{itemize}
\section{Related Works}
\subsection{Text-Based Visual Spatial Reasoning}
With rapid progress in embodied intelligence, spatial understanding in MLLMs has drawn increasing attention. Recent efforts to improve spatial reasoning in MLLMs can be broadly grouped into two categories. The first leverages curated datasets and multi-stage training to learn spatial relations from large-scale examples~\citep{chen2024spatialvlm, li2025spatialladder, yang2025visual, cai2025scaling, elmaaroufi2025graid, liu2025spatial, zhan2025actial, batra2025spatialthinker, ouyang2025spacer}. For example, SpatialLadder~\citep{li2025spatialladder} constructs SpatialLadder-26k and adopts a progressive curriculum from perception to reasoning. SpaceR~\citep{ouyang2025spacer} further improves spatial reasoning by introducing spatially tailored reward signals in reinforcement learning. However, direct training often promotes memorization over reasoning, limiting generalization.

Recognizing the limitations of purely 2D priors, the second category augments the input with 2.5D or 3D information to strengthen global spatial understanding~\citep{mao2025spatiallm, zheng2025video, huang2025mllms, wu2025spatial, zhang2025spatial, gholami2025spatial}. For example, SpatialMind~\citep{zhang2025spatial} encodes object locations on a 2D grid to form a cognitive map that is provided as additional input. Spatial-MLLM~\citep{wu2025spatial} incorporates a frozen geometry encoder (e.g. VGGT) to provide complementary 3D representations. Nevertheless, methods that primarily generate text still struggle to faithfully express continuous spatial relations in discrete language. This mismatch causes substantial information loss when continuous geometry is mapped to textual descriptions, ultimately limiting spatial understanding.

\subsection{Latent Reasoning}
To overcome the limitations of a text-only output modality, one straightforward solution is to ``think with images''~\citep{yang2025mindjourney, wu2024mind, wu2025reinforcing, gao2024cantor}. For example, ViLaSR~\citep{wu2025reinforcing} facilitates spatial localization by rendering auxiliary guide lines, while MindJourney~\citep{yang2025mindjourney} leverages a diffusion model to synthesize alternative viewpoints and feeds the generated images back into the model. However, these methods rely heavily on external tools, reducing flexibility and limiting generalization.

Inspired by latent reasoning in LLMs~\citep{butt2025soft, wei2025sim, shen2025codi}, which replaces discrete text tokens with self-generated continuous embeddings, several studies have extended latent reasoning to MLLMs~\citep{ray2025mull, qin2025chain, yang2025machine, li2025latent, wang2025monet} and further to spatial reasoning~\citep{bigverdi2025perception, liu2025ssr, hu2025g, chen2025think} to improve reasoning flexibility. LVR~\citep{li2025latent} and Mirage~\citep{yang2025machine} use image supervision to guide latent tokens, encouraging the model to develop an interleaved visual--textual reasoning process in latent space. For spatial reasoning, Aurora~\citep{bigverdi2025perception} and SSR~\citep{liu2025ssr} incorporate depth maps into latent tokens to strengthen geometric reconstruction. Nevertheless, the complexity of spatial reasoning is unlikely to be fully captured by a single global latent token. Such a bottleneck often lacks sufficient capacity and structure for diverse spatial problems, while also reducing interpretability when the model must resolve heterogeneous geometric relations.

\begin{figure*}[t]
\includegraphics[width=\linewidth]{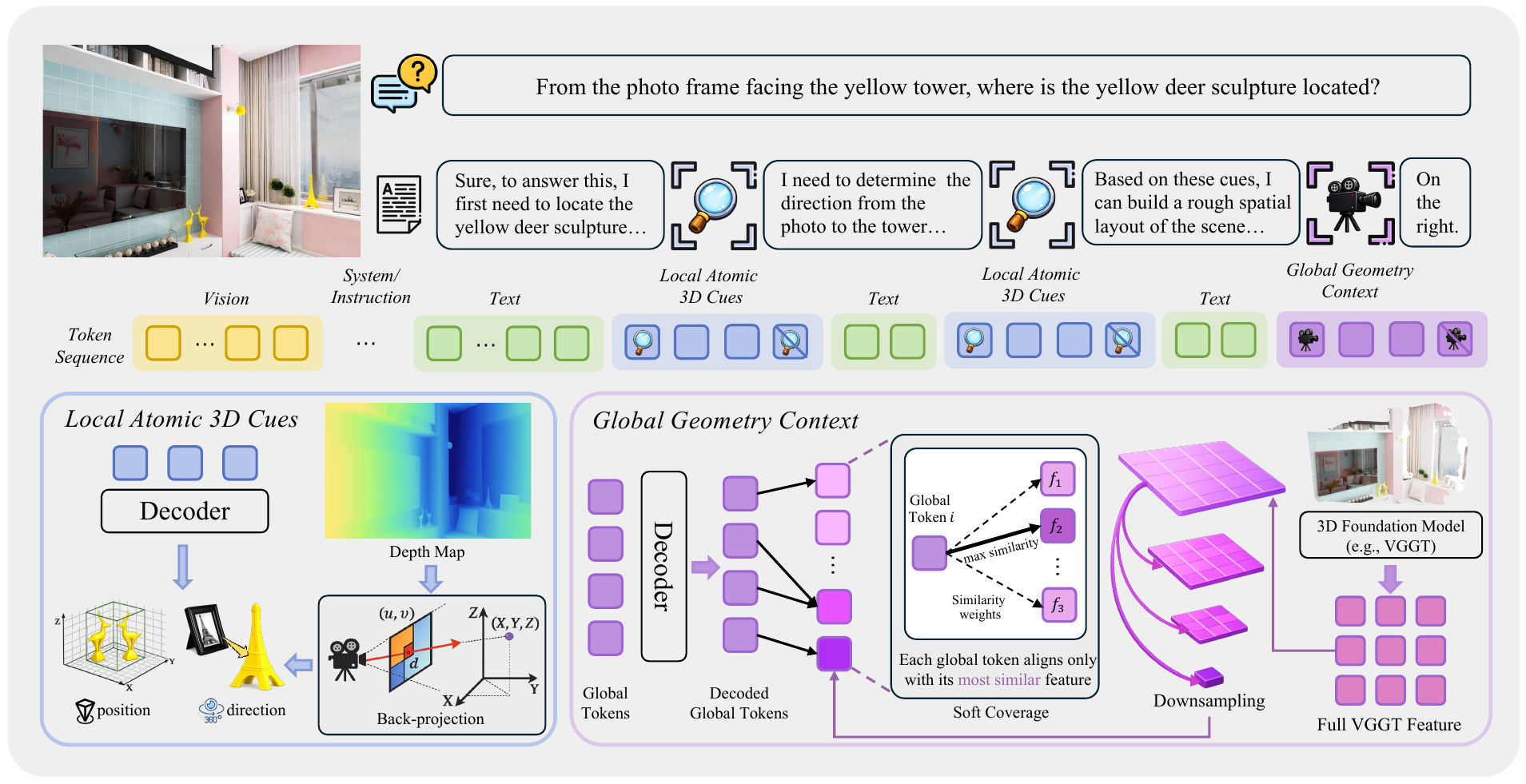}
\caption{\textbf{Overview of GeoAnchor.} We organize spatial reasoning into two components: (i) obtaining local atomic 3D cues and (ii) building global geometric context. We replace discrete text-based representations with continuous latent tokens to encode abstract 3D information. Specifically, local tokens capture object locations and inter-object direction, whereas the global token encodes the overall scene structure.}
\label{fig:oveall_structure}
\end{figure*}

\section{Method}

In this section, we present GeoAnchor, a framework designed to tackle complex spatial reasoning tasks via an interleaved text-latent paradigm. Section~\ref{sec:arch} presents the overall architecture, while Section~\ref{sec:latent} details the design of decomposed spatial tokens.  Section~\ref{sec:training} then describes the collaborative training strategy, which facilitates a structured evolution from local perception to holistic spatial understanding. Finally, Section~\ref{sec:dataset} summarizes the construction of the grounding and reasoning datasets employed across training stages.

\subsection{Text-Latent Interleaved Framework}\label{sec:arch}

GeoAnchor employs a text-latent interleaved reasoning framework to address the loss of continuous 3D information inherent in relying solely on discrete textual representations. 
Specifically, the model utilizes discrete text tokens for semantic planning and logical reasoning, while leveraging latent representations to encode and capture over continuous, multi-grained 3D spatial cues. 
The overall structure is illustrated in Figure~\ref{fig:oveall_structure}.
Given an image $\mathcal{I}$ and query $Q$, GeoAnchor outputs the reasoning trajectory $\mathcal{O}$, which consists of an interleaved sequence of text and latent tokens:

\begin{equation}
    \mathcal{O} = t_1 \oplus z_1 \oplus \cdots \oplus z_{k-1} \oplus t_k,
\end{equation}
where $\mathbf{t} = \{t_1, \ldots, t_k\}$ denotes the sequence of text tokens, and $\mathbf{z} = \{z_1, \ldots, z_{k-1}\}$ represents the sequence of latent tokens.

Unlike text tokens retrieved via embedding lookups, each latent token $z_i$ is defined as a fixed-length sequence of $M$ continuous hidden states, i.e., $z_i = \{h_{i,1}, h_{i,2}, \dots, h_{i,M}\}$, where each $h_{i,j} \in \mathbb{R}^d$ resides in the model's hidden space. Specifically, at the $j$-th internal step of generating $z_i$, the MLLM $f_\theta(\cdot)$ produces the hidden state $h_{i,j}$ conditioned on the preceding context:
\begin{equation}
    h_{i,j} = f_{\theta}^{\mathrm{hidden}}(Q, \mathcal{I}, \mathcal{O}_{<z_i}, h_{i,1:j-1}),
\end{equation}
where $\mathcal{O}_{<z_i}$ denotes the entire reasoning trajectory generated before $z_i$, and $h_{i,1:j-1}$ denotes the internal hidden states generated up to step $j-1$.
However, the output hidden-state space and the input embedding space typically lie on different manifolds. Directly feeding $h_{i,j}$ back into the input trajectory without alignment can lead to severe latent drift and autoregressive instability~\citep{yue2025hybrid}. To bridge this gap, we introduce a latent projector $\mathcal{P}(\cdot)$ that maps the output hidden state into the input embedding space. Specifically, the projected continuous embedding $e_{i,j} \in \mathbb{R}^d$ is computed as:
\begin{equation}
    e_{i,j} = \mathcal{P}(h_{i,j}) = \text{LayerNorm}\Big(h_{i,j} + \text{MLP}\big(\text{LayerNorm}(h_{i,j})\big)\Big).
\end{equation}
The projected embedding $e_{i,j}$ is then fed back as the input representation for the subsequent generation step, ensuring a consistent and stable reasoning trajectory.

\subsection{Decomposed Spatial Tokens}\label{sec:latent}
To address the limitation that a single latent type struggles to handle diverse spatial reasoning tasks, we decompose the spatial reasoning process into two distinct stages: obtaining local atomic 3D cues and building global geometric context. Accordingly, we introduce three complementary latent tokens: the position token $z^{\mathrm{pos}}$, the direction token $z^{\mathrm{dir}}$, and the geometry token $z^{\mathrm{geo}}$. The first two serve as local tokens by encoding atomic 3D cues for target objects, including object grounding and relational orientation, whereas the geometry token is a global token to capture the overall scene structure. By disentangling the uniform latent representation into structured 3D evidence with distinct functions, the model can flexibly compose diverse spatial evidence within the latent space, thereby enabling more interpretable spatial reasoning.

\noindent\textbf{Explicit Supervision of Local 3D Cues.}
In conventional text-based reasoning, local spatial information, such as positional coordinates and directional vectors, is typically represented as discrete numerical text tokens. To preserve the continuity of 3D spatial information, we encode these cues into latent tokens and employ decoders to map them into numerical coordinates for alignment.
We map latent tokens to the target 3D space using lightweight linear heads, similar to the text prediction module. Given that each latent token comprises a sequence of hidden states (Section~\ref{sec:arch}), $z^{\mathrm{pos}}$ and $z^{\mathrm{dir}}$ are defined as:
\begin{equation}
    z^{\mathrm{pos}}=\{\mathbf{h}^{\mathrm{pos}}_1,\ldots,\mathbf{h}^{\mathrm{pos}}_{l_{\mathrm{pos}}}\},\qquad
    z^{\mathrm{dir}}=\{\mathbf{h}^{\mathrm{dir}}_1,\ldots,\mathbf{h}^{\mathrm{dir}}_{l_{\mathrm{dir}}}\},
\end{equation}
where $\mathbf{h}^{\mathrm{pos}}_j$ and $\mathbf{h}^{\mathrm{dir}}_j\in\mathbb{R}^D$ are the hidden states at the $j$-th internal step, and $l_{\mathrm{pos}}$ and $l_{\mathrm{dir}}$ denote the numbers of hidden states for each latent token. The hidden states within each latent token are first averaged along the sequence dimension and then mapped to 3D outputs through two separate linear heads:
\begin{equation}
    \hat{\mathbf{p}}=W_{\mathrm{pos}}\frac{1}{l_{\mathrm{pos}}}\sum_{j=1}^{l_{\mathrm{pos}}}\mathbf{h}^{\mathrm{pos}}_j, \quad
    \hat{\mathbf{d}}=W_{\mathrm{dir}}\frac{1}{l_{\mathrm{dir}}}\sum_{j=1}^{l_{\mathrm{dir}}}\mathbf{h}^{\mathrm{dir}}_j,
\end{equation}
where $W_{\mathrm{pos}}, W_{\mathrm{dir}} \in \mathbb{R}^{3 \times D}$ are projection heads, while $\hat{\mathbf{p}} \in \mathbb{R}^3$ and $\hat{\mathbf{d}} \in \mathbb{R}^3$ denote the predicted position and direction, respectively.
To explicitly supervise these two tokens, we apply the Smooth L1 loss~\citep{girshick2015fast} for position prediction and the cosine similarity loss for direction prediction:
\begin{equation}
    \mathcal{L}_{\mathrm{pos}} = 
    \begin{cases} 
      \frac{1}{2}(\hat{\mathbf{p}} - \mathbf{p})^2, & \text{if } |\hat{\mathbf{p}} - \mathbf{p}| < 1.0 \\
      |\hat{\mathbf{p}} - \mathbf{p}| - \frac{1}{2}, & \text{otherwise}
    \end{cases},
    \mathcal{L}_{\mathrm{dir}} = 1 - \frac{\hat{\mathbf{d}} \cdot \mathbf{d}}{\|\hat{\mathbf{d}}\|_2 \|\mathbf{d}\|_2},
\end{equation}
where $\mathbf{p}$ and $\mathbf{d}$ denote the ground-truth 3D position coordinates and direction vectors, respectively.

\noindent\textbf{Coverage-Based Supervision of Global Geometry.}
We supervise the global token $z^{\mathrm{geo}}$ using geometry features from the last layer of the VGGT model~\citep{wang2025vggt}, since these features can be decoded into 3D point clouds and inherently capture global geometric structure.
However, VGGT generates a high-resolution feature grid, whereas the global token is a compact representation composed of a few hidden states. 
Consequently, enforcing strict token-wise alignment would impose overly fine-grained constraints on the global token, potentially undermining its role as a compact representation of scene-level geometry. Therefore, a coarse soft-coverage alignment strategy is adopted instead of strict dense alignment.

We first transform the dense VGGT feature map into a supervision feature by multi-scale average pooling. Specifically, for the final VGGT feature map of size $H\times W\times D$, average pooling is applied at multiple spatial resolutions $\{r_1,\dots,r_L\}$. 
Each pooled feature grid is flattened into $r_l^2$ feature vectors, and the vectors from all scales are concatenated to form the coarse-grained VGGT feature map $V \in \mathbb{R}^{l_{\mathrm{vggt}} \times D}$, where $l_{\mathrm{vggt}} = \sum_{l=1}^{L} r_l^2$.
In parallel, the global token $z^{\mathrm{geo}}$ is projected into geometry tokens $G \in \mathbb{R}^{l_{\mathrm{geo}} \times D}$ via a linear layer, where $l_{\mathrm{geo}}$ denotes the number of geometry tokens. We then compute token similarities as:
\begin{equation}
A_{i,j}=\frac{\langle \mathbf{v}_{i}, \mathbf{g}_{j}\rangle}{\tau},
\qquad
u_j=\frac{1}{l_{\mathrm{vggt}}}\sum_{i=1}^{l_{\mathrm{vggt}}}\frac{\exp(A_{i,j})}{\sum_{j'=1}^{l_{\mathrm{geo}}}\exp(A_{i,j'})},
\end{equation}
where $\mathbf{v}_i \in \mathbb{R}^D$ is the $i$-th coarse-grained VGGT feature and $\mathbf{g}_j \in \mathbb{R}^D$ is the $j$-th geometry token, $\tau$ represents the temperature parameter, and $u_j$ denotes the average soft assignment received by the $j$-th geometry token. The alignment loss is defined as:
\begin{equation}
\mathcal{L}_{\mathrm{geo}}
=
-\frac{1}{l_{\mathrm{vggt}}}\sum_{i=1}^{l_{\mathrm{vggt}}}\log\sum_{j=1}^{l_{\mathrm{geo}}}\exp(A_{i,j})
+
\lambda_{\mathrm{bal}}
\sum_{j=1}^{l_{\mathrm{geo}}}u_j\left(\log u_j-\log\frac{1}{l_{\mathrm{geo}}}\right).
\end{equation}
The first term encourages that every VGGT feature aligns with at least one geometry token, thereby achieving coarse coverage without dense alignment. The second term enforces balanced token utilization to prevent representation collapse, controlled by the coefficient $\lambda_{\mathrm{bal}}$. Collectively, these objectives guide the geometry tokens to capture complementary global evidence that integrates seamlessly with local evidence for downstream reasoning.

\begin{figure*}[t]
\includegraphics[width=\linewidth]{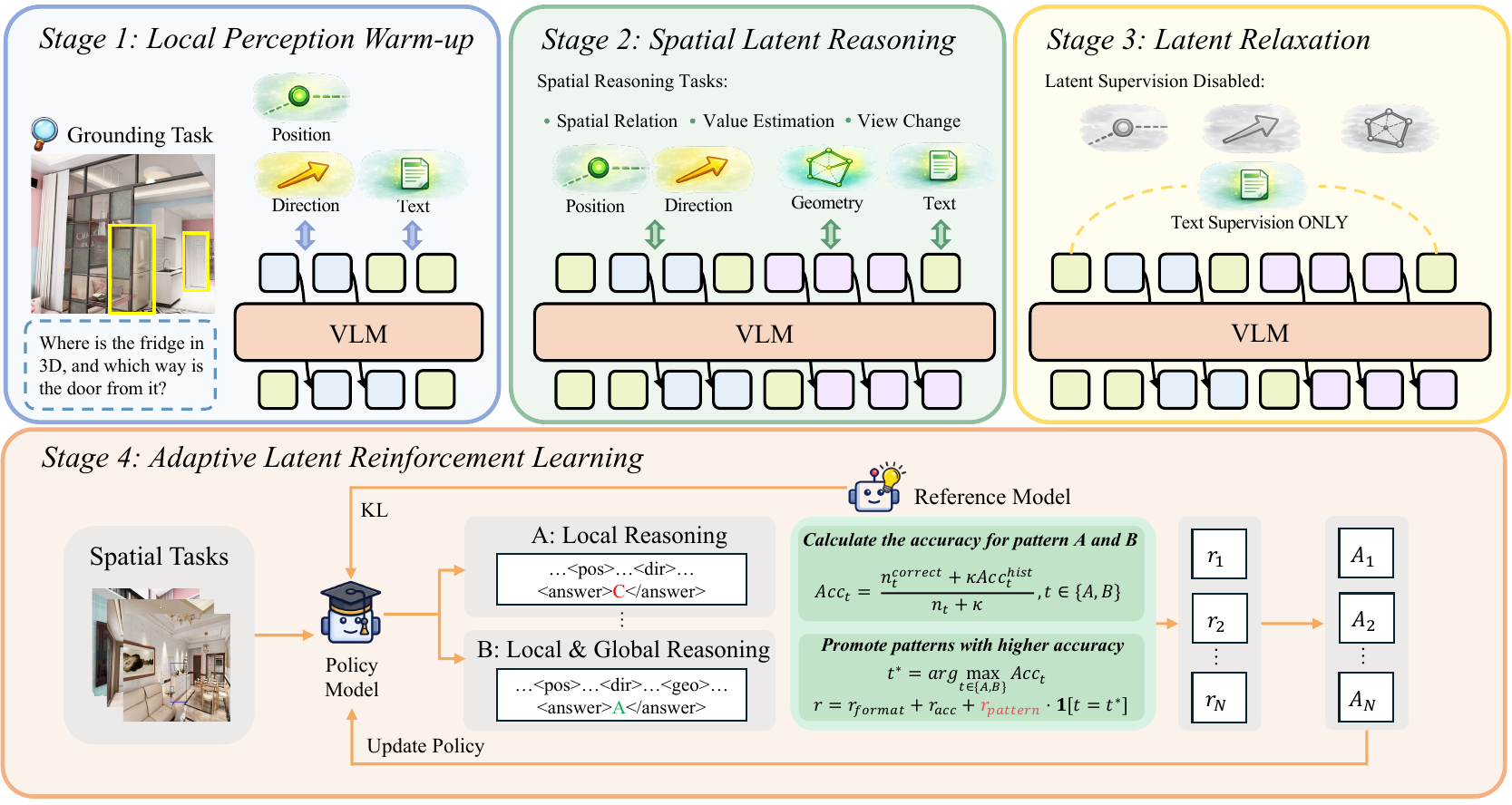}
\caption{\textbf{Overview of the collaborative training framework of GeoAnchor.} The model is trained in a coarse-to-fine manner, evolving from local perception to holistic spatial understanding, then to latent relaxation without explicit latent supervision, and finally to an adaptive latent reinforcement learning stage that encourages dynamic reasoning patterns.}
\label{fig:training}
\end{figure*}

\subsection{Collaborative Training Strategy}\label{sec:training}
To effectively train the model for spatial reasoning with latent tokens, we design a collaborative training framework that systematically builds up spatial intelligence, as illustrated in Figure~\ref{fig:training}.

\noindent\textbf{Stage 1: Local Perception Warm-up.} We first warm up the base model by training its 3D grounding capability. Specifically, we construct a large-scale 3D grounding dataset and train the model with two complementary abilities: predicting the 3D position of a target object and predicting the relative direction between two objects. By learning these two signals jointly, the model is encouraged to collaboratively organize the local tokens into coherent local spatial evidence. 
The local tokens are optimized with $\mathcal{L}_{\text{pos}}$ and $\mathcal{L}_{\text{dir}}$, while the text tokens are trained with the next-token prediction loss $\mathcal{L}_{\text{NTP}}$. Formally, the training objective for Stage~1 is:
\begin{equation}
    \mathcal{L}_{\mathrm{stage1}} = \lambda_t \mathcal{L}_{\text{NTP}} + \lambda_l \left( \mathcal{L}_{\mathrm{pos}} + \mathcal{L}_{\mathrm{dir}} \right),
\end{equation}
where $\lambda_t$ and $\lambda_l$ are the corresponding balance coefficients.


\noindent\textbf{Stage 2: Spatial Latent Reasoning.} In the second stage, the model is further trained on a spatial reasoning dataset covering diverse spatial tasks. During this stage, it learns to conduct an interpretable spatial reasoning process in the latent space by collaboratively integrating local 3D cues with global geometric context. It also learns to dynamically utilize the position and the direction tokens according to the input question, thereby developing a basic ability for adaptive latent token selection. The position, direction, and geometry tokens are supervised by their corresponding losses. Denoting $\lambda_g$ as the coefficient for $\mathcal{L}_{\mathrm{geo}}$, the total loss for Stage~2 is defined as follows: 
\begin{equation}
    \mathcal{L}_{\mathrm{stage2}}=\lambda_t\mathcal{L}_{\text{NTP}} + \lambda_l(\mathcal{L}_{\mathrm{pos}} + \mathcal{L}_{\mathrm{dir}}) + \lambda_g\mathcal{L}_{\mathrm{geo}}.
\end{equation}

\noindent\textbf{Stage 3: Latent Relaxation.}
Although the explicit supervision in Stage~2 effectively aligns the latent tokens with 3D information, overly strong alignment may shift them away from the original language manifold and hinder subsequent reasoning. Inspired by~\citep{yang2025machine, ray2025mull}, we therefore introduce a latent relaxation stage that removes explicit latent supervision and retains only text supervision. $\lambda_{l}$ and $\lambda_{g}$ are set to 0, and the latent tokens are thus updated only through the text supervision, which helps the learned spatial representations better fit the language distribution while preserving the spatial information acquired in earlier stages. 

\noindent\textbf{Stage 4: Adaptive Latent Reinforcement Learning.}
Although the preceding stages enable reasoning with latent tokens, they still rely on a fixed usage pattern where both local and global tokens are invoked for every query. However, different spatial reasoning tasks require different levels of spatial information, making such fixed invocation suboptimal.
To address this issue, we introduce an adaptive latent reinforcement learning stage based on Group Relative Policy Optimization (GRPO)~\citep{shao2024deepseekmath}. We consider two reasoning patterns: local-only and local-plus-global.
Then in addition to format and accuracy rewards, we introduce a pattern-specific reward to evaluate the effectiveness of each reasoning pattern.
Within each group, the model samples $N$ responses, each generated by randomly selecting one of the two patterns. For pattern $t \in \{\text{local}, \text{local+global}\}$, its effectiveness is estimated as: 
\begin{equation}
Acc_t = \frac{n_t^{\mathrm{correct}} + \kappa Acc_t^{\mathrm{hist}}}{n_t + \kappa},
Acc_t^{\mathrm{hist}} \leftarrow (1-\mu)\, Acc_t^{\mathrm{hist}} + \mu \frac{n_t^{\mathrm{correct}}}{n_t},
\end{equation}
where $n_t$ and $n_t^{\mathrm{correct}}$ denote the numbers of sampled and correct responses under pattern $t$, respectively, $Acc_t^{\mathrm{hist}}$ is the EMA-based historical accuracy~\citep{kingma2014adam}, $\kappa$ is the smoothing coefficient, and $\mu$ is the update rate. The pattern with the higher smoothed accuracy receives an additional pattern reward $r_{\text{pattern}}$. This encourages the model to use global token only when local reasoning is insufficient, leading to more efficient and more interpretable reasoning.

\subsection{Dataset Construction}\label{sec:dataset}
\newcommand{\best}[1]{\textbf{#1}}
\newcommand{\second}[1]{\underline{#1}}
\newcommand{\modelcell}[2]{#1~{\scriptsize[#2]}}

\definecolor{impgreen}{RGB}{255, 0, 0}

\begin{table*}[t]
  \centering
  \caption{Evaluation Results on Spatial Reasoning Benchmarks. For each metric, \textbf{bold} numbers indicate the best performance, while \underline{underlined} numbers represent the second-best performance.}
  \label{tab:benchmark_results}

  \resizebox{\textwidth}{!}{%
  \begin{tabular}{lc *{12}{c}}
    \toprule
    & & \multicolumn{6}{c}{\textbf{SPAR-Bench}} & \multicolumn{3}{c}{\textbf{SPBench}} & \multicolumn{3}{c}{\textbf{ViewSpatial}} \\
    \cmidrule(lr){3-8}\cmidrule(lr){9-11}\cmidrule(lr){12-14}
    \textbf{Model} & \textbf{Param}
      & \textbf{Avg.} & \textbf{Dep.} & \textbf{Dis.} & \textbf{Prox.} & \textbf{Rel.} & \textbf{View}
      & \textbf{Avg.} & \textbf{Rel.} & \textbf{Abs.}
      & \textbf{Avg.} & \textbf{Cam.} & \textbf{Per.} \\
    \midrule

    \multicolumn{14}{l}{\textbf{\textit{Proprietary Models}}}\\
    GPT-4o~\citep{hurst2024gpt} & -
      & 40.1 & 34.3 & 43.4 & 57.1 & 45.9 & 31.0
      & 53.4 & 49.4 & 56.0
      & 37.5 & 33.5 & 43.6 \\
    Gemini-2.5-Flash~\citep{comanici2025gemini} & -
      & \second{48.7} & 37.4 & 45.2 & \second{80.1} & 63.2 & \second{41.0}
      & 51.5 & 44.7 & 56.0
      & 44.0 & 43.0 & \second{45.5} \\
    \midrule

    \multicolumn{14}{l}{\textbf{\textit{General Models}}}\\
    MiniCPM-V-4.5~\citep{yu2025minicpm} & 8B
      & 37.7 & 32.4 & 32.6 & 62.1 & 50.6 & 29.8
      & 40.9 & 47.4 & 36.7
      & 39.0 & 43.0 & 32.9 \\
    LLaVA-OneVision-1.5~\citep{an2025llava} & 8B
      & 34.5 & 29.3 & 32.6 & 58.5 & 40.7 & 26.7
      & 40.8 & 45.1 & 38.0
      & 38.7 & 40.6 & 35.3 \\
    Qwen3-VL~\citep{yang2025qwen3} & 8B
      & 41.1 & 32.5 & 34.6 & 71.2 & 62.1 & 31.5
      & 55.1 & 47.6 & \second{60.0}
      & 44.0 & 44.6 & 43.1 \\
    Molmo2~\citep{clark2026molmo2} & 8B
      & 24.6 & 16.7 & 3.2 & 62.4 & 48.4 & 25.1
      & 29.7 & 47.6 & 18.1
      & 45.5 & 46.0 & 45.1 \\
    GLM-4.1V~\citep{hong2025glm} & 9B
      & 48.0 & \second{42.4} & 53.2 & 65.6 & 57.7 & 34.3
      & 49.1 & 43.6 & 52.6
      & 40.4 & 37.6 & 44.7 \\
    Qwen3.5~\citep{qwen3.5} & 9B
      & 48.6 & 39.0 & 53.0 & 73.8 & \second{63.5} & 33.4
      & 48.8 & 48.6 & 48.8
      & \second{45.6} & \second{47.2} & 43.2 \\
    Kimi-VL~\citep{team2025kimi} & 16B-A3B
      & 33.9 & 28.6 & 26.7 & 60.6 & 44.5 & 28.3
      & 42.1 & 40.8 & 43.0
      & 38.1 & 36.5 & 40.4 \\
    Internvl3.5~\citep{wang2025internvl3} & 38B
      & 38.0 & 31.5 & 35.5 & 60.9 & 58.0 & 25.5
      & 53.1 & 49.4 & 55.5
      & 41.4 & 43.3 & 38.6 \\
    \midrule

    \multicolumn{14}{l}{\textbf{\textit{Specialized Models}}}\\
    \modelcell{G2VLM~\citep{hu2025g}}{CVPR'26} & 2B
      & 41.7 & 41.8 & \second{66.3} & 37.7 & 26.4 & 24.2
      & 14.3 & 25.2 & 7.2
      & 18.3 & 16.0 & 21.7 \\
    \modelcell{3DThinker~\citep{chen2025think}}{CVPR'26} & 3B
      & 23.8 & 15.8 & 18.6 & 49.7 & 24.2 & 25.1
      & 13.8 & 32.2 & 1.7
      & 29.2 & 27.7 & 31.5 \\
    \modelcell{SpatialLadder~\citep{li2025spatialladder}}{ICLR'26} & 3B
      & 32.4 & 22.1 & 29.8 & 53.8 & 43.7 & 29.5
      & \second{69.6} & \second{81.6} & \best{65.0}
      & 43.3 & 41.8 & 45.4 \\
    \modelcell{SpatialMLLM-v1.1~\citep{wu2025spatial}}{NeurIPS'25} & 4B
      & 33.7 & 25.3 & 34.2 & 53.8 & 36.5 & 31.2
      & 54.2 & 55.7 & 53.3
      & 39.7 & 39.8 & 39.6 \\
    \modelcell{ViLaSR~\citep{wu2025reinforcing}}{NeurIPS'25} & 7B
      & 38.6 & 28.9 & 43.1 & 61.8 & 46.4 & 28.2
      & 52.2 & 54.4 & 50.8
      & 34.3 & 33.8 & 34.9 \\
    \modelcell{Aurora~\citep{bigverdi2025perception}}{CVPR'25} & 13B
      & 29.6 & 26.9 & 29.0 & 52.7 & 21.7 & 25.7
      & 32.1 & 36.8 & 29.0
      & 25.6 & 26.2 & 24.7 \\
    \midrule
    Qwen3-VL~\citep{yang2025qwen3} (Baseline Model) & 2B
      & 32.4 & 21.9 & 24.0 & 60.6 & 49.2 & 28.3
      & 52.9 & 51.4 & 54.8
      & 36.3 & 39.5 & 37.2 \\
    \textbf{GeoAnchor} & 2B
      & \best{68.4} & \best{55.5} & \best{67.4} & \best{80.3} & \best{84.6} & \best{68.8}
      & \best{69.7} & \best{86.7} & 58.7
      & \best{47.0} & \best{47.5} & \best{46.0} \\
    \textcolor{impgreen}{\textit{Improvement}} & -
      & \textcolor{impgreen}{+36.0} & \textcolor{impgreen}{+33.6} & \textcolor{impgreen}{+43.4} & \textcolor{impgreen}{+19.7} & \textcolor{impgreen}{+35.4} & \textcolor{impgreen}{+40.5}
      & \textcolor{impgreen}{+16.8} & \textcolor{impgreen}{+35.3} & \textcolor{impgreen}{+3.9}
      & \textcolor{impgreen}{+10.7} & \textcolor{impgreen}{+8.0} & \textcolor{impgreen}{+8.8} \\
    \bottomrule
  \end{tabular}%
  }
\end{table*}
We construct two training sets for GeoAnchor: a large-scale 3D grounding dataset for Stage 1 and a spatial reasoning dataset for the subsequent supervised and reinforcement learning stages. The former provides explicit supervision for local spatial perception, while the latter supports joint local-global spatial reasoning.

For the grounding dataset, we build on top of ScanNet~\citep{dai2017scannet} by sampling 10k scenes from its 2.5 million views over 1,500+ scans. For each image, we use Qwen3-VL-32B~\citep{yang2025qwen3} to identify up to five objects, along with their descriptions and 2D bounding boxes. We then employ Depth Anything v3~\citep{lin2025depth} to estimate depth maps and camera poses, and recover the 3D coordinates of the extracted objects through back-projection. Ground-truth directions are derived from coordinate differences. We further diversify the queries using point-coordinate, bounding-box, and natural-language formulations, with each sample involving one to three objects. This process yields 550k 3D grounding samples in total.

For the spatial reasoning dataset, we use SPAR~\citep{zhang2025flatland} as the primary data source, since it is built on ScanNet~\citep{dai2017scannet}, ScanNet++~\citep{yeshwanth2023scannet++}, and Structured3D~\citep{zheng2020structured3d}, and provides 7M samples covering diverse spatial tasks. We sample 100k questions from SPAR and use the bounding boxes provided in the dataset to obtain the ground-truth positions and directions of the target objects, following the same back-projection procedure as in the grounding dataset. In addition, we extract global geometric features for each image using VGGT~\citep{wang2025vggt}. To better cover centimeter-based answer formats, we further incorporate 5k samples from SpatialLadder-26k~\citep{li2025spatialladder}. In total, the resulting spatial reasoning dataset contains 105k samples.
\section{Experiment}
\subsection{Experimental Setup}\label{sec:setting}
\noindent\textbf{Implementation Details.} 
GeoAnchor is built on Qwen3-VL-2B~\citep{yang2025qwen3}, with local token length $l_{\mathrm{pos}}=l_{\mathrm{dir}}=2$ and global token length $l_{\mathrm{geo}}=8$. Stage~1 is trained on the grounding dataset for one epoch using a batch size of 64 and a learning rate of $1\times10^{-4}$. Stages~2 and~3 are each trained for one epoch on the spatial reasoning dataset with a batch size of 32 and a learning rate of $2\times10^{-5}$. The loss weights are set to $\lambda_t=\lambda_l=1$ and $\lambda_g=0.1$. For global geometry supervision, the VGGT feature map is pooled at $L=3$, with $\{r_1,r_2,r_3\}=\{1,2,4\}$. In the RL stage, 8 rollouts are performed per question with a sampling temperature of 1.0, a pattern reward $r_{\text{pattern}}$ of 0.5, a KL-divergence coefficient $\beta$ of 0.01, and a learning rate of $5\times10^{-7}$. Additional details are provided in the supplementary material.

\noindent\textbf{Evaluation Benchmarks and Metrics.}
We evaluate GeoAnchor on SPAR-Bench~\citep{zhang2025flatland}, SPBench~\citep{li2025spatialladder}, and ViewSpatial~\citep{li2025viewspatial}, where the first two serve as in-domain benchmarks and the last assesses out-of-domain generalization. For multiple-choice questions, we directly judge correctness and report the mean accuracy. For numerical questions, the mean accuracy is computed as the average accuracy across confidence thresholds from 0.5 to 0.9 with a step size of 0.05. 


\subsection{Main Results}
Table~\ref{tab:benchmark_results} compares GeoAnchor with state-of-the-art models across three benchmarks. Despite its compact 2B parameter size, GeoAnchor outperforms proprietary, open-source, and specialized MLLMs, including those with up to 38B parameters. On in-domain benchmarks, it achieves state-of-the-art mean accuracy on both SPAR-Bench (68.4\%) and SPBench (69.7\%), surpassing the base Qwen3-VL-2B by 36.0\% and 16.8\%, respectively. While SpatialLadder slightly outperforms GeoAnchor on the SPBench Abs. metric, its inconsistent performance across other benchmarks suggests a limitation in generalization. Notably, on the out-of-domain ViewSpatial benchmark, GeoAnchor attains 47.0\% mean accuracy, ranking first in both sub-tasks. The substantial 10.7\% gain over the base model demonstrates our method's robust generalization capability beyond the training distribution.

\subsection{Ablation Experiments}\label{sec:ablation}
\begin{figure}[t]
    \centering

    \begin{minipage}[t]{0.52\textwidth}
        \vspace{0pt}
        \begin{minipage}[t][4.5\baselineskip][t]{\linewidth}
            \captionof{table}{Ablation study on the effectiveness of latent reasoning. Vanilla SFT trains the model directly on question-answer pairs, and text CoT follows a ``localize-then-reason'' paradigm and is trained entirely in the text modality.}
            \label{tab:ablation_latent}
        \end{minipage}

        \scriptsize
        \setlength{\tabcolsep}{2.2pt}
        \renewcommand{\arraystretch}{1.05}
        \begin{adjustbox}{max width=\linewidth}
            \begin{tabular}{@{}lcccc@{}}
                \toprule
                \textbf{Model} & \textbf{SPAR} & \textbf{SPBench} & \textbf{ViewSpatial} & \textbf{Avg.} \\
                \midrule
                Qwen3-VL-2B~\citep{yang2025qwen3} & 32.4 & 52.9 & 36.3 & 40.5 \\
                \quad + vanilla SFT & 56.4 & 58.1 & 40.9 & 51.8 \\
                \quad + text CoT SFT & 60.1 & 52.9 & 42.2 & 51.7 \\
                Latent Reasoning~\citep{bigverdi2025perception} & 62.4 & 57.8 & 40.5 & 53.6 \\
                \midrule
                local tokens & 65.8 & 66.9 & 44.4 & 59.0 \\
                global token & 63.8 & 64.3 & 46.0 & 58.0 \\
                local + global token & \textbf{67.5} & \textbf{68.8} & \textbf{46.3} & \textbf{60.9} \\
                \bottomrule
            \end{tabular}
        \end{adjustbox}
    \end{minipage}
    \hfill
    \begin{minipage}[t]{0.46\textwidth}
        \vspace{0pt}
        \begin{minipage}[t][4.5\baselineskip][t]{\linewidth}
            \captionof{table}{Ablation results on the collaborative training strategy. S1, S2, and S3 represent the three-stage SFT, and S4 represents the RL stage.}
            \label{tab:unified_ablation}
        \end{minipage}

        \scriptsize
        \setlength{\tabcolsep}{1.5pt}
        \renewcommand{\arraystretch}{1.05}
        \begin{adjustbox}{max width=\linewidth}
            \begin{tabular}{@{}ccccccccc@{}}
                \toprule
                \multirow{2}{*}{\textbf{S1}} & \multirow{2}{*}{\textbf{S2}} & \multirow{2}{*}{\textbf{S3}} & \multicolumn{2}{c}{\textbf{S4}} & \multirow{2}{*}{\textbf{SPAR}} & \multirow{2}{*}{\textbf{SPBench}} & \multirow{2}{*}{\textbf{ViewSpatial}} & \multirow{2}{*}{\textbf{Avg.}} \\
                \cmidrule(lr){4-5}
                & & & \textbf{w/o $r_{\mathrm{pattern}}$} & \textbf{w/ $r_{\mathrm{pattern}}$} & & & & \\
                \midrule
                       & \cmark & \cmark &        &        & 63.1 & 61.9 & 42.7 & 55.9 \\
                \cmark &        & \cmark &        &        & 56.5 & 48.4 & 33.9 & 46.3 \\
                \cmark & \cmark &        &        &        & 52.8 & 41.0 & 19.8 & 37.9 \\
                \cmark & \cmark & \cmark &        &        & 67.5 & 68.8 & 46.3 & 60.9 \\
                \cmark & \cmark & \cmark & \cmark &        & 67.8 & 69.2 & 46.2 & 61.1 \\
                \cmark & \cmark & \cmark &        & \cmark & \textbf{68.4} & \textbf{69.7} & \textbf{47.0} & \textbf{61.7} \\
                \bottomrule
            \end{tabular}
        \end{adjustbox}
    \end{minipage}
\end{figure}

\begin{table}[t]
    \centering

    \begin{minipage}[t]{0.40\textwidth}
        \vspace{0pt}
        \captionof{table}{Ablation of local token interpretability across position, direction, and mixed questions.}
        \label{tab:local_token_interpretability}

        \scriptsize
        \renewcommand{\arraystretch}{1.38}
        \begin{tabular*}{\linewidth}{@{\extracolsep{\fill}}lccc@{}}
            \toprule
            \textbf{Problem Setting} & \textbf{Position} & \textbf{Direction} & \textbf{Mixed} \\
            \midrule
            Position token only & 65.2 & 36.5 & 55.8 \\
            Direction token only & 62.9 & 38.7 & 53.5 \\
            Full local tokens & \textbf{66.4} & \textbf{39.4} & \textbf{57.0} \\
            \bottomrule
        \end{tabular*}
    \end{minipage}
    \hfill
    \begin{minipage}[t]{0.58\textwidth}
        \vspace{0pt}
        \captionof{table}{Ablation of VGGT alignment strategies for supervising the global geometry token.}
        \label{tab:vggt_ablation}

        \scriptsize
        \renewcommand{\arraystretch}{1.05}
        \begin{tabular*}{\linewidth}{@{\extracolsep{\fill}}llcccc@{}}
            \toprule
            \textbf{Alignment} & \textbf{Sampling} & \textbf{SPAR} & \textbf{SPBench} & \textbf{ViewSpatial} & \textbf{Avg.} \\
            \midrule
            \multirow{3}{*}{Dense} & Mean Pooling & 59.9 & 66.8 & 45.5 & 57.4 \\
            & Adaptive Pooling & 66.9 & 66.5 & 44.4 & 59.3 \\
            & Linear Interpolation & 53.7 & 60.2 & 37.3 & 37.8 \\
            \midrule
            Soft Coverage & Multi-Scale Pooling & \textbf{68.4} & \textbf{69.7} & \textbf{47.0} & \textbf{61.7} \\
            \bottomrule
        \end{tabular*}
    \end{minipage}
\end{table}

\textbf{Effects of Different Reasoning Paradigms.} 
Table~\ref{tab:ablation_latent} compares the performance of different reasoning paradigms on spatial tasks. Compared with vanilla SFT, text CoT SFT does not yield consistent gains across benchmarks, suggesting that spatial information is difficult to incorporate effectively through textual reasoning. Therefore, the reasoning process is more easily influenced by language patterns than by geometric evidence. In contrast, by encoding continuous 3D information into latent tokens, GeoAnchor outperforms both vanilla SFT and text CoT SFT, indicating that latent reasoning provides a more effective way to leverage spatial information. Furthermore, GeoAnchor also surpasses the single-latent reasoning baseline. Performance drops when the model is restricted to one type of analysis, showing that the decomposed token design plays a more important role than latent reasoning. Our design reduces the semantic burden of a single latent and enables more task-adaptive use of spatial cues, leading to more interpretable and robust reasoning than a single-latent design.

\begin{figure}[t]
    \centering
    \includegraphics[width=\linewidth]{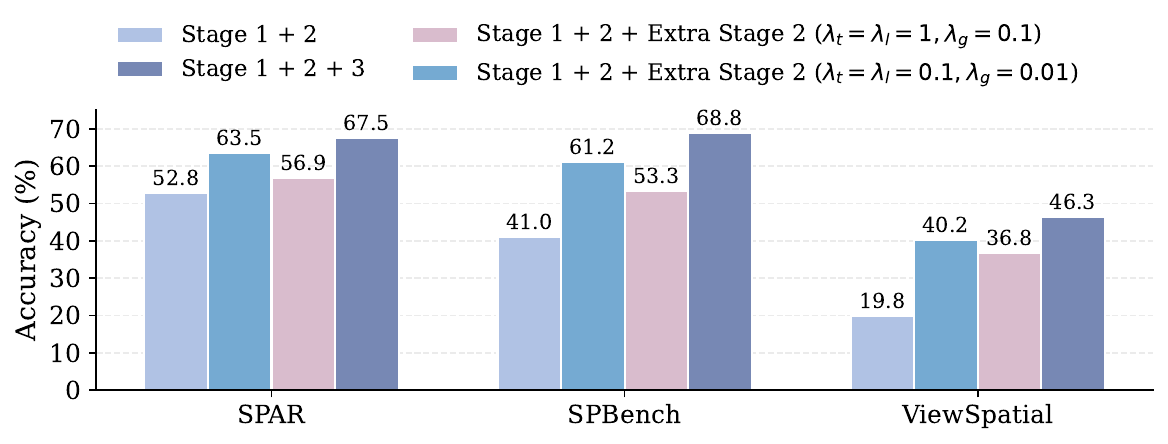}
    \caption{Ablation study on the effectiveness of Stage 3.}
    \label{fig:stage3_ablation}
\end{figure}

\noindent \textbf{Effects of Collaborative Training Stages.} 
Table~\ref{tab:unified_ablation} illustrates the effectiveness of each training stage. Stages 1, 2, and 3 improve the performance by 5.0\%, 14.6\%, and 23.0\%, respectively, validating the effectiveness of our multi-stage design.
Figure~\ref{fig:stage3_ablation} shows that Stage 3 yields significantly higher gains than merely extending Stage 2, confirming its value lies in integrating latent tokens into reasoning rather than additional training iterations. 
Moreover, while applying GRPO directly as stage 4 improves in-domain performance, it degrades out-of-domain results on ViewSpatial. 
After adding the pattern reward, the model achieves consistent gains across all three benchmarks. This suggests that applying a fixed reasoning pattern introduces redundant information and thereby interferes with answer prediction. In contrast, the pattern reward encourages a more concise and efficient reasoning process, enabling the model to make more accurate use of spatial evidence.

\noindent \textbf{Effects of Local Position and Direction Tokens.} Table~\ref{tab:local_token_interpretability} groups the three benchmarks into three categories: Position, which focuses on relative object locations; Direction, which focuses on object orientation; and Mixed, which requires both types of local evidence. The results show that using both position and direction tokens yields the best overall performance. When only the position token is used during reasoning, the model performs well on position-related questions but shows clear degradation on direction-related and mixed categories. Using only the direction token leads to the opposite trend. These observations indicate that the local tokens indeed encode their corresponding spatial information and provide strong evidence for final answer generation.

\begin{figure*}[t]
    \centering
    \begin{subfigure}[t]{0.32\textwidth}
        \centering
        \includegraphics[width=\linewidth]{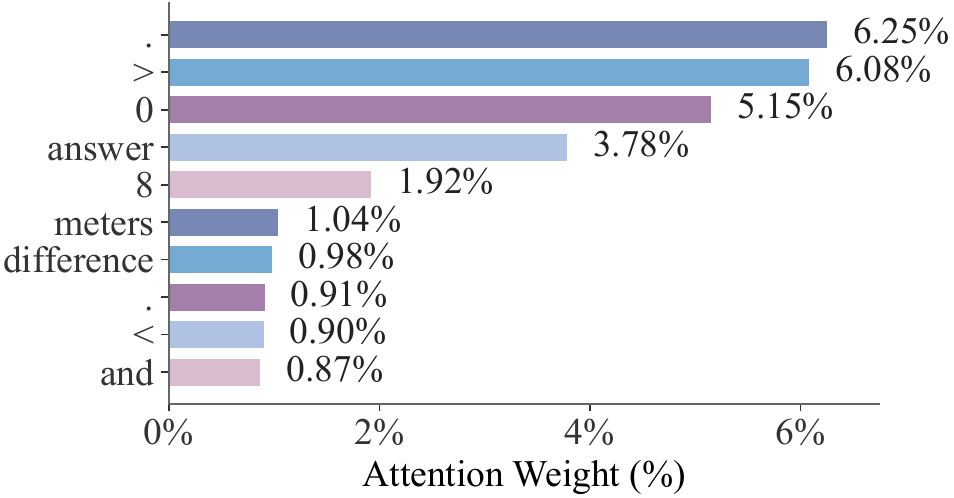}
        \caption{Text CoT}
        \label{fig:text_cot}
    \end{subfigure}
    \hfill
    \begin{subfigure}[t]{0.32\textwidth}
        \centering
        \includegraphics[width=\linewidth]{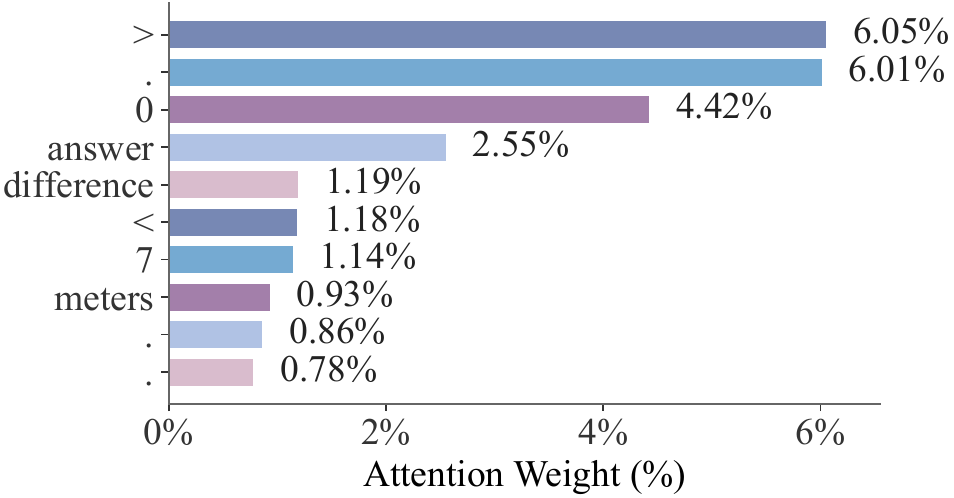}
        \caption{GeoAnchor w/o Stage 1}
        \label{fig:GeoAnchor_wo_1}
    \end{subfigure}
    \hfill
    \begin{subfigure}[t]{0.32\textwidth}
        \centering
        \includegraphics[width=\linewidth]{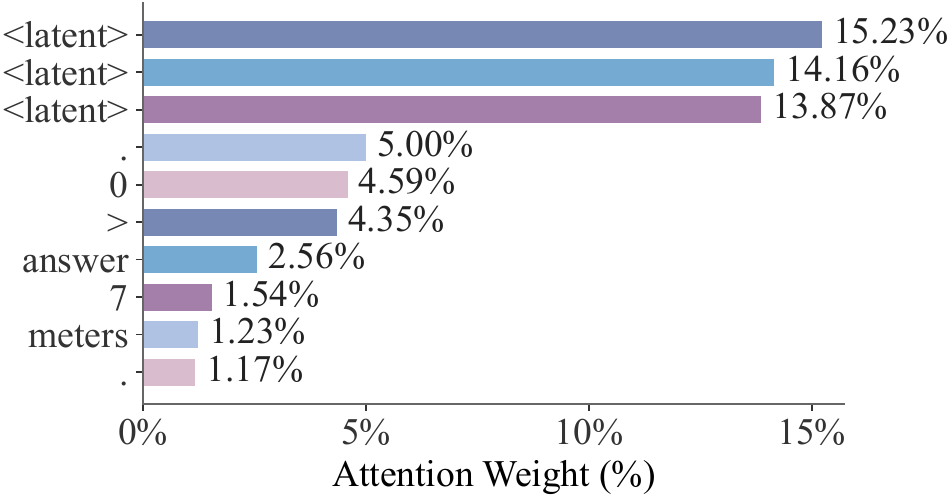}
        \caption{GeoAnchor}
        \label{fig:GeoAnchor}
    \end{subfigure}
    \caption{Comparison of the top text attention weights assigned by the final answer under different reasoning paradigms.}
    \label{fig:text_attention}
\end{figure*}

\begin{figure}[t]
    \centering

    \begin{minipage}[t]{0.48\linewidth}
        \vspace{0pt}
        \centering
        \includegraphics[width=\linewidth]{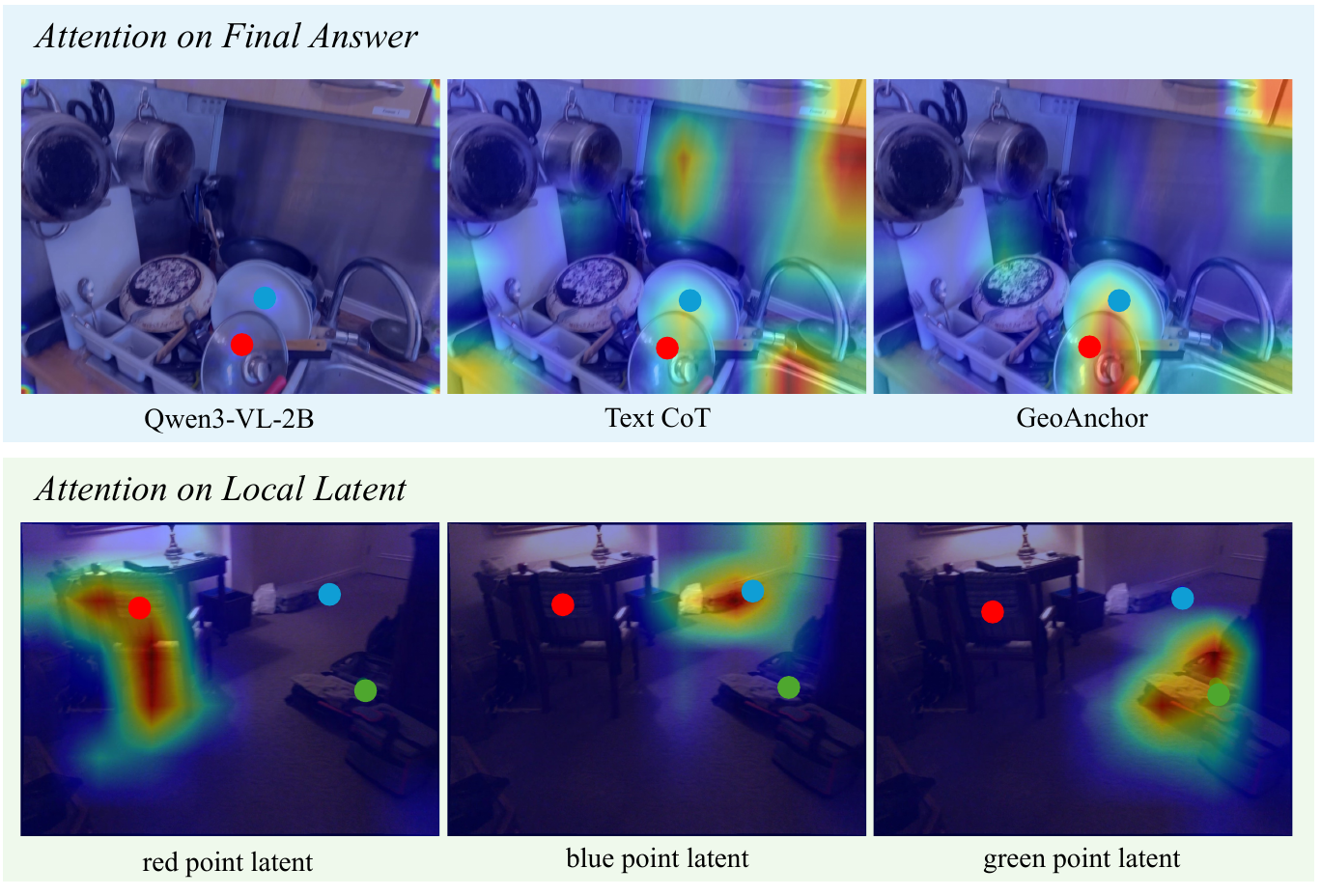}

        \captionsetup{skip=3pt}
        \captionof{figure}{Visual attention comparison on the final answer and local token. The red, blue, and green points represent the target objects mentioned in the questions.}
        \label{fig:image_attention}
    \end{minipage}
    \hfill
    \begin{minipage}[t]{0.50\linewidth}
        \vspace{0pt}
        \centering
        \includegraphics[width=\linewidth]{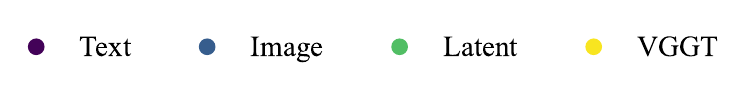}

        \vspace{3pt}

        \begin{minipage}[t]{0.49\linewidth}
            \centering
            \includegraphics[width=\linewidth]{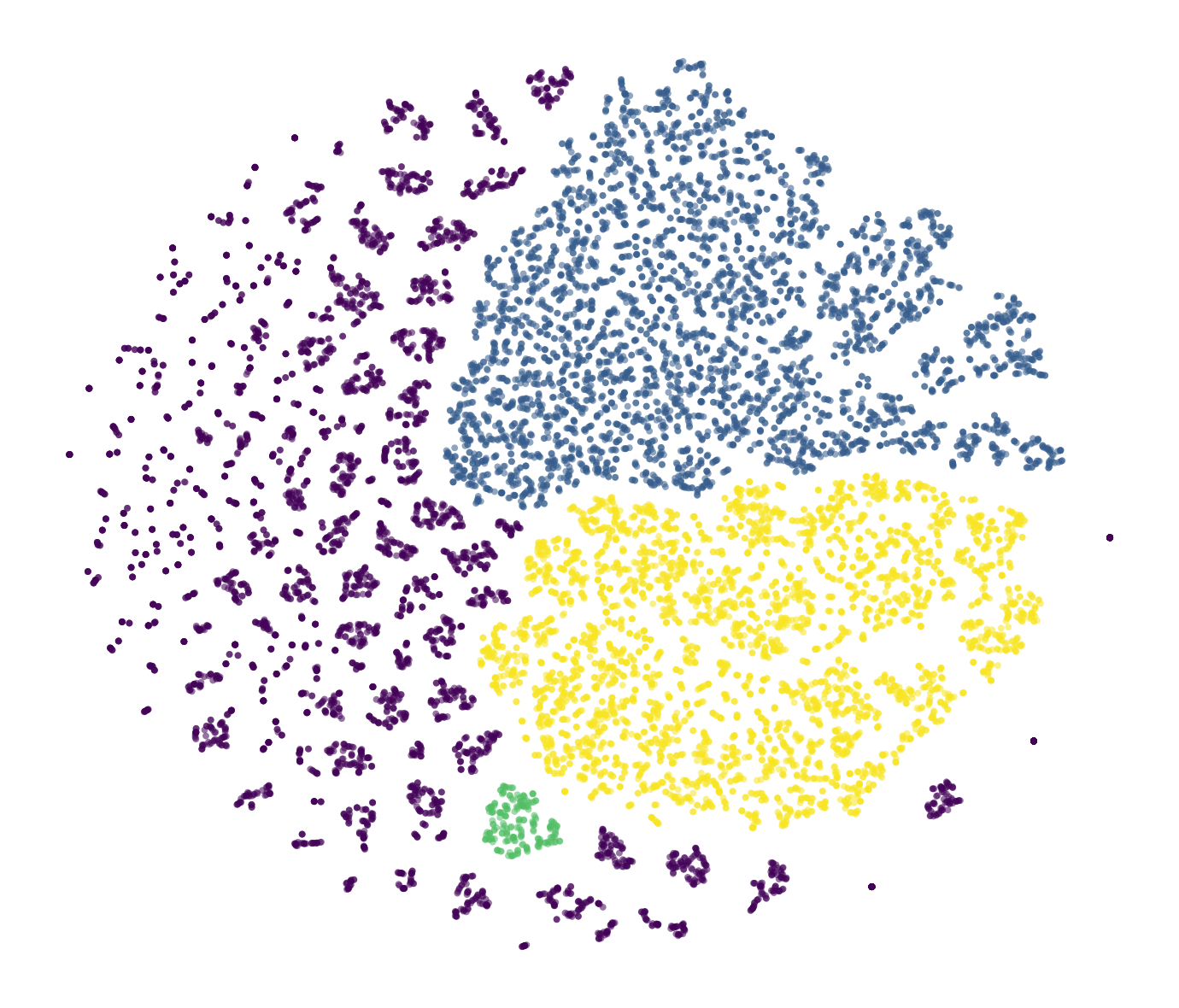}\\[-2pt]
            \small SPAR-Bench
        \end{minipage}
        \hfill
        \begin{minipage}[t]{0.49\linewidth}
            \centering
            \includegraphics[width=\linewidth]{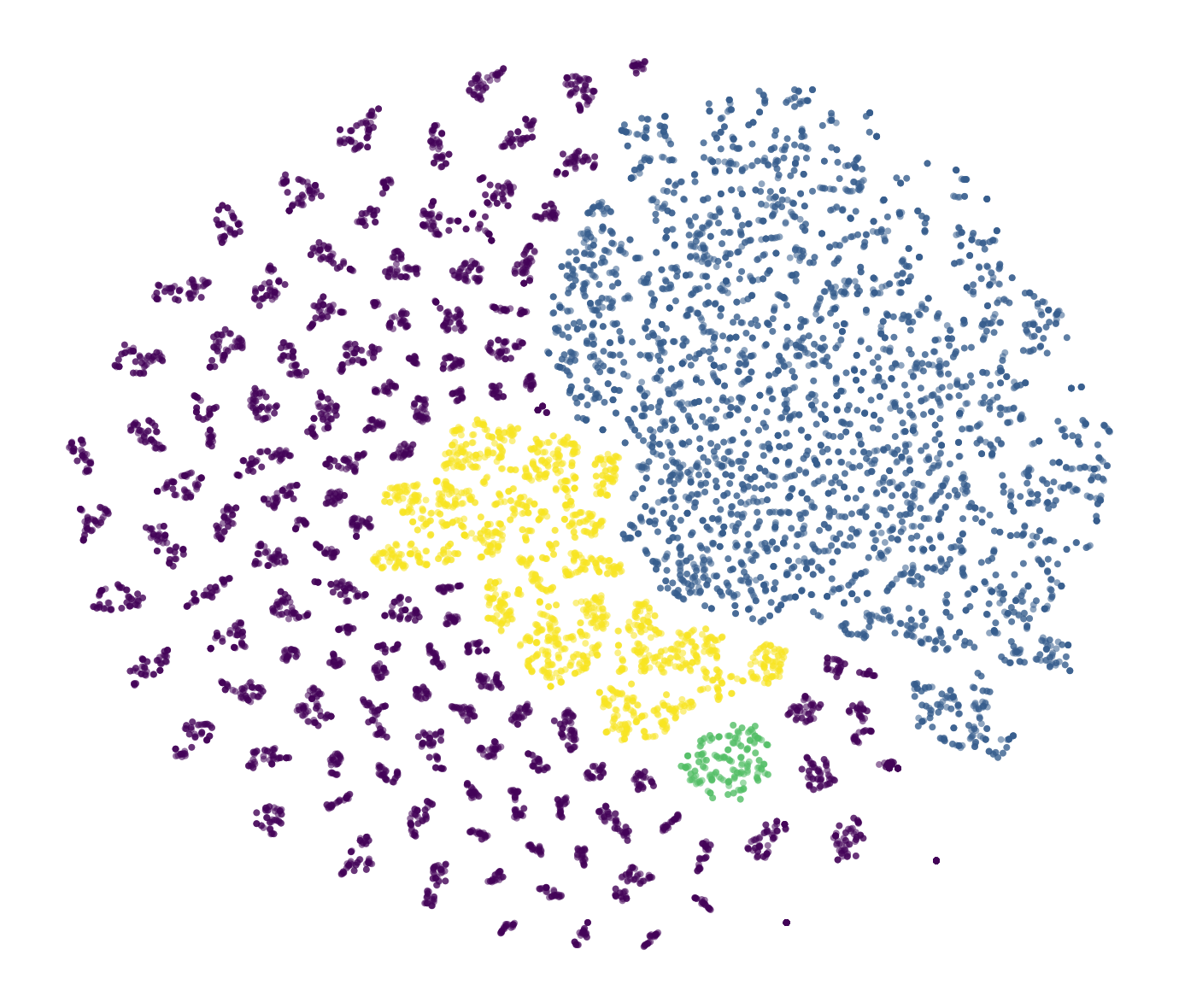}\\[-2pt]
            \small SPBench
        \end{minipage}

        \captionsetup{skip=16pt}
        \captionof{figure}{Visualization of text, image, VGGT, and global token embeddings using t-SNE.}
        \label{fig:tsne_visual}
    \end{minipage}
\end{figure}

\noindent \textbf{Effects of Alignment Strategies in Global Tokens.} Table~\ref{tab:vggt_ablation} compares different VGGT alignment strategies. The proposed soft coverage alignment is evaluated against three dense alignment baselines. Specifically, \textit{Mean Pooling} averages the VGGT features and the global token along the sequence dimension. \textit{Adaptive Pooling} and \textit{Linear Interpolation} resample the VGGT features to match the length of the global token using average pooling and interpolation, respectively. 
The results show that soft coverage alignment achieves the best performance across all three benchmarks, suggesting that coverage-based supervision provides a more suitable strategy for learning a compact representation of scene-level geometry while preserving sufficient information for downstream reasoning.

\subsection{In-Depth Analysis}\label{sec:analysis}

\textbf{Latent reasoning improves aggregation of answer-relevant information, with Stage 1 playing a key role.}
As shown in Figure~\ref{fig:text_attention}, GeoAnchor assigns stronger attention to latent tokens when generating the final answer. In contrast, although Text CoT also produces numerical coordinates, its answer token attends less to spatially meaningful cues and more to semantically weak symbols. This suggests that discretized textual reasoning weakens spatial semantics, whereas GeoAnchor allows the answer to directly leverage compact latent representations. The attention analysis further highlights that Stage 1 is important for developing a strong dependence on latent tokens, indicating that this stage establishes robust grounded latent evidence that supports subsequent reasoning.

\noindent\textbf{Final-answer attention is more accurately grounded in task-relevant visual regions.}
As shown in the upper part of Figure~\ref{fig:image_attention}, GeoAnchor focuses more precisely on the visual regions relevant to the queried spatial target when generating the final answer. By comparison, the baseline exhibits more diffuse attention, while Text CoT spreads attention more broadly without clearly isolating the most relevant object or location. This suggests that GeoAnchor not only aggregates reasoning information more effectively, but also grounds its predictions more faithfully in visual evidence.

\noindent\textbf{Local tokens show clear spatial specialization.}
The lower part of Figure~\ref{fig:image_attention} shows that GeoAnchor produces distinct visual attention patterns when generating latent tokens for different positions. Each local token consistently attends to the region associated with its corresponding spatial target, indicating that these tokens encode disentangled spatial semantics rather than entangled global patterns. This behavior suggests that the latent space is organized in a semantically meaningful way, making the intermediate reasoning variables both interpretable and useful for final answer prediction.

\noindent\textbf{The global token indeed encodes VGGT information.} 
Following \cite{yang2025machine}, we use t-SNE~\citep{van2008visualizing} to visualize the relationships among text, image, VGGT features, and the global token. As shown in Figure~\ref{fig:tsne_visual}, on both SPAR-Bench and SPBench, the global token lies between the VGGT and text tokens. This suggests that it remains close to the geometry manifold while preserving textual semantics, consistent with the role of Stage 3 discussed in Section~\ref{sec:training}.




\section{Conclusion}
This paper presents GeoAnchor, a text-latent interleaved framework for spatial reasoning. By decomposing spatial reasoning into position, direction, and geometry latents, GeoAnchor enables decomposed local-plus-global reasoning beyond text-only representations. A collaborative multi-stage training strategy further improves the learning of grounded spatial evidence. Experiments on multiple benchmarks show that GeoAnchor consistently outperforms the base model and generalizes well across diverse 3D reasoning tasks.

\section{Acknowledgement}
This work is supported by the National Natural Science Foundation of China (Grant Nos. 92270201 and 62125204), and National Natural Science Foundation of China (62522102, 62432001), Beijing Natural Science Foundation (L247006).

\newpage
\bibliographystyle{plainnat}
\bibliography{paper}

\appendix
\clearpage
\setcounter{page}{1}
\setcounter{table}{0}
\setcounter{figure}{0}
\setcounter{equation}{0}
\section{Datasets and Evaluation Benchmarks}
Our training data mainly come from SPAR-7M and SpatialLadder-26K. In the experiments, we use SPAR-Bench, SPBench, and ViewSpatial-Bench as the main benchmarks to evaluate our method. In this section, we first introduce the datasets and benchmarks used in our study, and then summarize the task definitions and sample counts of the evaluated benchmark subsets in the corresponding tables.

\subsection{SPAR-7M and SPAR-Bench}
SPAR-7M is a large-scale dataset designed for spatial understanding, constructed from more than 4,000 indoor 3D scenes collected from public scene datasets with 3D ground truth, including ScanNet, ScanNet++, and Structured3D. Through a 3D-driven data generation pipeline, the authors derive image sequences, camera parameters, and depth information from these scenes, and further generate diverse spatial question-answer pairs from the associated 3D scene metadata. As a result, SPAR-7M covers 33 task types and contains over 7 million QA pairs, spanning a broad range of abilities from low-level spatial perception to high-level spatial reasoning, and supporting single-view, multi-view, and video settings.

Importantly, the dataset is accompanied by its own benchmark, namely SPAR-Bench, which is explicitly introduced for evaluation rather than training alone. SPAR-Bench is constructed from the SPAR-7M split by selecting representative spatial tasks and manually verifying the resulting samples for quality control. In our experiments, we use the single-image subset of SPAR-Bench, which contains 2,866 samples in total. Detailed task definitions and per-task sample counts are summarized in Table~\ref{tab:spar_tasks}.
\begin{table}[h]
\centering
\caption{Task definitions and sample counts of the SPAR-Bench subset used in our experiments.}
\label{tab:spar_tasks}
\small
\setlength{\tabcolsep}{3pt}
\begin{tabularx}{\columnwidth}{>{\raggedright\arraybackslash}p{4.0cm} >{\raggedright\arraybackslash}X c}
\toprule
\textbf{Task Name} & \textbf{Description} & \textbf{\#Samples} \\
\midrule
depth\_prediction\_oc & Given the depth of a reference point, estimate the depth of another queried point. & 360 \\
depth\_prediction\_oo & Given the depth of a reference point, estimate the depth difference between two other queried points. & 372 \\
distance\_prediction\_oc & Estimate the distance between a queried point and the camera. & 393 \\
distance\_prediction\_oo & Estimate the distance between two queried points. & 363 \\
distance\_infer\_center\_oo & Determine which of two queried points is farther away from a given reference point. & 340 \\
obj\_spatial\_relation & Under the camera view, determine the relative position of one queried point with respect to another. & 364 \\
spatial\_imagination\_oc & First determine the position of a queried point under the camera view, then re-judge its position after a viewpoint transformation. & 372 \\
spatial\_imagination\_oo & First determine the relative position between two queried points under the camera view, then re-judge their relation after a viewpoint transformation. & 302 \\
\midrule
\textbf{Total} & & \textbf{2866} \\
\bottomrule
\end{tabularx}
\end{table}

\subsection{SpatialLadder-26K and SPBench}
SpatialLadder-26K is a multimodal spatial reasoning dataset designed to support progressive learning from perceptual grounding to more advanced reasoning. It contains 26,610 samples across four complementary task categories, including object localization, single-image spatial reasoning, multi-view spatial reasoning, and video spatial reasoning. In terms of task coverage, the dataset spans seven spatial dimensions, namely relative direction, relative distance, absolute distance, object size, counting, room size, and appearance order, thereby covering a broad range of spatial understanding skills across image, multi-view, and video modalities. In terms of data sources, the object localization, single-image, and multi-view portions are constructed from ScanNet 3D scene reconstructions, while the video subset is sampled from SR-91k.

The paper further introduces two dedicated benchmarks built using the same pipeline on the ScanNet validation set, namely SPBench-SI and SPBench-MV. In our experiments, we use only the single-image benchmark, namely SPBench-SI, which contains 1,009 samples in total. Detailed task definitions and per-task sample counts are summarized in Table~\ref{tab:spbench_tasks}.

\begin{table}[h]
\centering
\caption{Task definitions and sample counts of the SPBench-SI subset used in our experiments.}
\label{tab:spbench_tasks}
\small
\setlength{\tabcolsep}{3pt}
\begin{tabularx}{\columnwidth}{>{\raggedright\arraybackslash}p{4cm} >{\raggedright\arraybackslash}X c}
\toprule
\textbf{Task Name} & \textbf{Description} & \textbf{\#Samples} \\
\midrule
object\_rel\_direction & Under the camera view, determine the relative position between two objects. & 306 \\
object\_rel\_distance & Determine which object is farther away from a given reference object. & 91 \\
object\_abs\_distance & Under the camera view, estimate the distance between two objects. & 149 \\
object\_size\_estimation & Estimate the size of an object along a given dimension in centimeters. & 463 \\
\midrule
\textbf{Total} & & \textbf{1009} \\
\bottomrule
\end{tabularx}
\end{table}

\subsection{ViewSpatial-Bench}
ViewSpatial-Bench is a dedicated benchmark for evaluating multi-perspective spatial localization in vision-language models. It contains over 5,700 multiple-choice question-answer pairs spanning more than 1,000 unique 3D scenes, with source images drawn from the validation sets of ScanNet and MS-COCO. In terms of task coverage, the benchmark comprises multiple task types under complementary perspective settings, targeting both camera-perspective and person-perspective spatial reasoning. This design makes it a challenging benchmark for evaluating cross-viewpoint spatial understanding.

Although the paper also describes a larger multi-perspective spatial training dataset generated by the same automated 3D annotation pipeline, the associated training set is not publicly released. Accordingly, in our experiments, we use only the single-image evaluation subset of ViewSpatial-Bench, which contains 4,607 samples in total. Detailed task definitions and per-task sample counts are summarized in Table~\ref{tab:viewspatial_tasks}.
\begin{table}[t]
\centering
\caption{Task definitions and sample counts of the ViewSpatial-Bench subset used in our experiments.}
\label{tab:viewspatial_tasks}
\small
\setlength{\tabcolsep}{3pt}
\begin{tabularx}{\columnwidth}{>{\raggedright\arraybackslash}p{3.5cm} >{\raggedright\arraybackslash}X c}
\toprule
\textbf{Task Name} & \textbf{Description} & \textbf{\#Samples} \\
\midrule
Camera perspective -- Relative Direction & Under the camera view, determine the relative position of one queried point with respect to another. & 1773 \\
Camera perspective -- Object View Orientation & Given an image, determine the facing orientation of the object. & 996 \\
Person perspective -- Object View Orientation & Determine the facing orientation of the person appearing in the image. & 996 \\
Person perspective -- Relative Direction & Given a specified person-centered perspective, determine the relative position of an object. & 842 \\
\midrule
\textbf{Total} & & \textbf{4607} \\
\bottomrule
\end{tabularx}
\end{table}

\subsection{Definition of Grouped Metrics}
Main results in the paper using grouped abbreviations instead of listing every original task separately. Table~\ref{tab:metric_grouping} summarizes the mapping from each grouped metric to its corresponding original tasks. In all cases, each grouped metric is computed as the average over the corresponding original tasks.

\begin{table}[t]
\centering
\caption{Mapping from grouped metric abbreviations to the original tasks.}
\label{tab:metric_grouping}
\small
\begin{tabular}{l|ll}
\toprule
\textbf{Benchmark} & \textbf{Abbr.} & \textbf{Original Task} \\
\midrule

\multirow{5}{*}{SPAR-Bench}
& Dep.  & depth\_prediction\_oc, depth\_prediction\_oo \\
& Dis.  & distance\_prediction\_oc, distance\_prediction\_oo \\
& Prox.                  & distance\_infer\_center\_oo \\
& Rel.                   & obj\_spatial\_relation \\
& View. & spatial\_imagination\_oc, spatial\_imagination\_oo \\
\midrule

\multirow{2}{*}{SPBench}
& Rel. & object\_rel\_direction, object\_rel\_distance \\
& Abs. & object\_abs\_distance, object\_size\_estimation \\
\midrule

\multirow{4}{*}{ViewSpatial}
& \multirow{2}{*}{Cam.} & Camera perspective -- Relative Direction \\
&                       & Camera perspective -- Object View Orientation \\
& \multirow{2}{*}{Per.} & Person perspective -- Object View Orientation \\
&                       & Person perspective -- Relative Direction \\
\bottomrule
\end{tabular}
\end{table}

\section{Training Dataset Construction}
\subsection{Overview}
We construct two training datasets for GeoAnchor. The first is a 3D grounding dataset used in Stage 1 to initialize local spatial perception. The second is a spatial reasoning training dataset used in Stages 2 and 3 as well as the subsequent RL stage. In this section, we describe the construction pipeline, annotation procedure, and summary statistics of these datasets.

\subsection{3D Grounding Dataset}
\subsubsection{Object Extraction}
\begin{figure*}[t]
    \centering
    \begin{minipage}[t]{0.49\textwidth}
        \vspace{0pt}
        \centering
        \includegraphics[width=\linewidth]{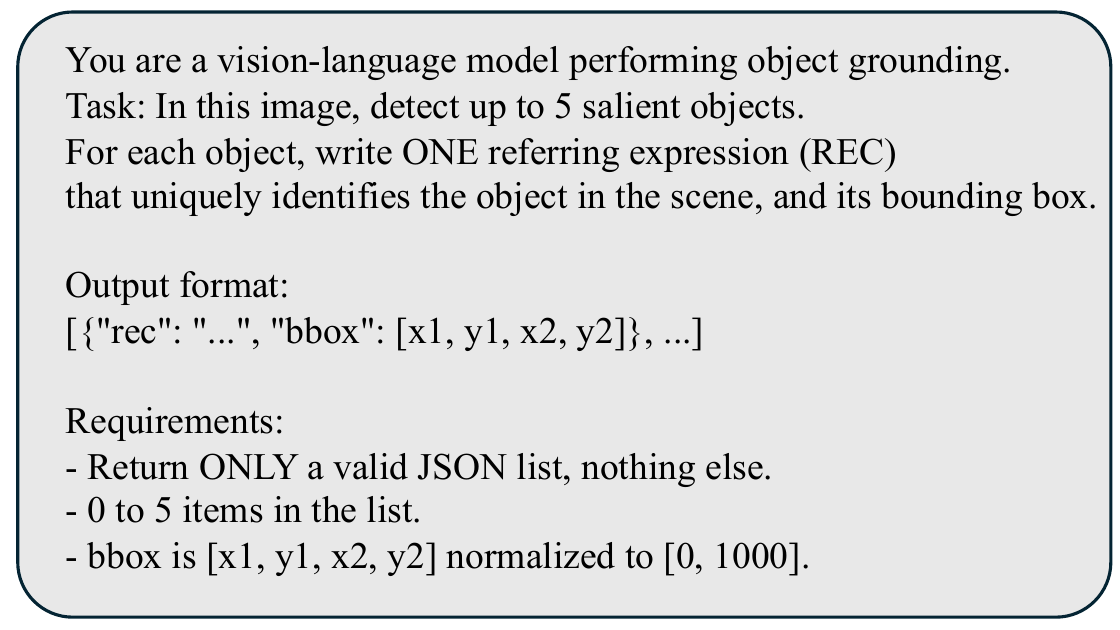}
        \captionof{figure}{Prompt used for object extraction and referring-expression generation in the construction pipeline of the 3D grounding dataset.}
        \label{fig:grounding_prompt}
    \end{minipage}
    \hfill
    \begin{minipage}[t]{0.49\textwidth}
        \vspace{0pt}
        \centering
        \includegraphics[width=\linewidth]{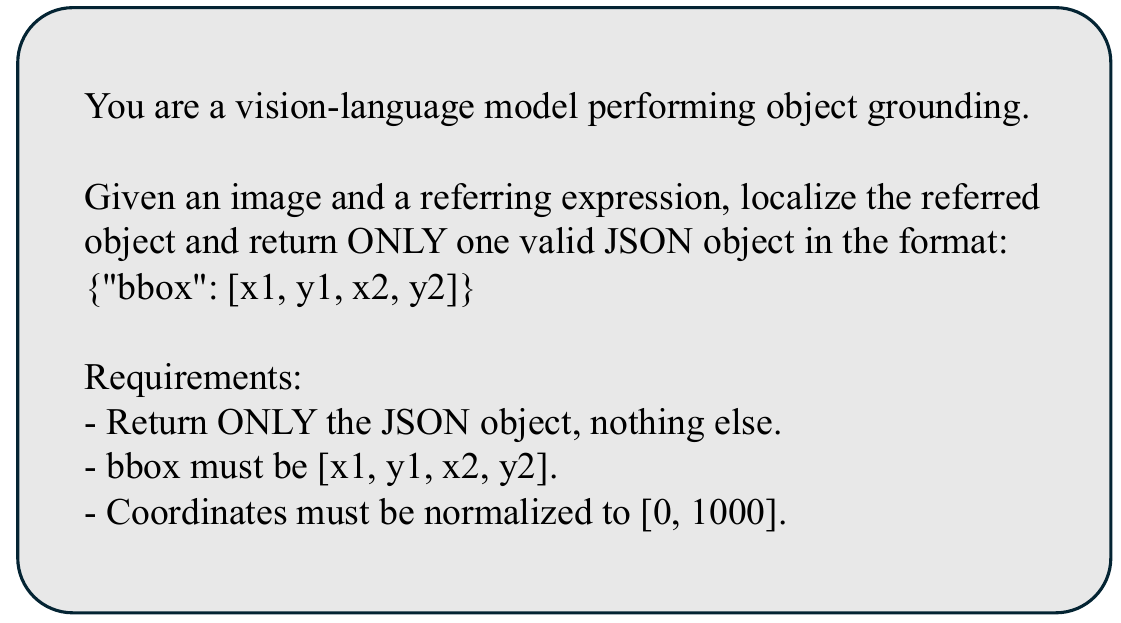}
        \captionof{figure}{Prompt used for the self-consistency verification of automatically generated referring expression--bounding box pairs.}
        \label{fig:verification_prompt}
    \end{minipage}
\end{figure*}

Our 3D grounding dataset is constructed based on ScanNet. We sample 10k scenes from more than 2.5 million views spanning over 1,500 scans. For each image, we first employ Qwen3-VL-32B to identify up to five salient objects, together with their textual descriptions and 2D bounding boxes. The prompt used for object extraction is shown in Figure.~\ref{fig:grounding_prompt}. 

To ensure the consistency between the generated referring expressions and the localized regions, we perform an additional verification step after object extraction. Concretely, for each object, we take the generated referring expression and use it as a grounding query on the same image to obtain a second predicted bounding box. We then measure the overlap between the original box and the re-grounded box using intersection-over-union (IoU). An annotation is retained only if the IoU exceeds 0.5; otherwise, it is discarded. This self-consistency check effectively filters out noisy cases in which the generated textual description does not faithfully correspond to the extracted object region. The prompt used for verification is shown in Figure.~\ref{fig:verification_prompt}. 

Table~\ref{tab:grounding_verification_stats} reports the verification results. Across 10,000 images, 47,663 object annotations are automatically extracted, and 44,998 pass the self-consistency check, giving an object-level retention rate of 94.41\%. The mean and median IoU between the original and re-grounded boxes are 0.787 and 0.856, showing that most generated referring expression--bounding box pairs are well aligned.

\begin{table*}[t]
    \centering

    \begin{minipage}[t]{0.49\textwidth}
        \vspace{0pt}
        \centering
        \captionof{table}{Statistics of the self-consistency verification step for object extraction.}
        \label{tab:grounding_verification_stats}
        
        \resizebox{\linewidth}{!}{%
        \begin{tabular}{ccccc}
            \toprule
            \textbf{Total anns.}
            & \textbf{Verified anns.}
            & \textbf{Retention rate}
            & \textbf{Mean IoU}
            & \textbf{Median IoU} \\
            \midrule
            47,663 & 44,998 & 94.41\% & 0.787 & 0.856 \\
            \bottomrule
        \end{tabular}%
        }
    \end{minipage}
    \hfill
    \begin{minipage}[t]{0.49\textwidth}
        \vspace{0pt}
        \centering
        \captionof{table}{Comparison between 3D object coordinates recovered from pseudo depth and ground-truth depth.}
        \label{tab:pseudo_depth_3d}
        
        \resizebox{\linewidth}{!}{%
        \begin{tabular}{ccccccc}
            \toprule
            \textbf{Mean Err.}
            & \textbf{Median Err.}
            & \textbf{$x$ Err.}
            & \textbf{$y$ Err.}
            & \textbf{$z$ Err.}
            & \textbf{Acc@0.5m}
            & \textbf{Acc@0.2m} \\
            \midrule
            0.09 & 0.04 & 0.02 & 0.01 & 0.08 & 98.0 & 92.9 \\
            \bottomrule
        \end{tabular}%
        }
    \end{minipage}

\end{table*}

\begin{table}[t]
\centering
\caption{Statistics of 3D Grounding Dataset used in Stage 1.}
\label{tab:grounding_dataset}
\begin{tabular}{lc}
\toprule
\textbf{Question Type} & \textbf{Count} \\
\midrule
Single Position Question   & 224,990 \\
Multiple Position Question & 159,530 \\
Direction Question         & 164,920 \\
\midrule
\textbf{Total}             & \textbf{549,440} \\
\bottomrule
\end{tabular}
\end{table}

\subsubsection{3D Position Acquisition}

We estimate depth maps and camera poses using Depth Anything v3, and recover object 3D coordinates through depth-guided back-projection. These recovered coordinates are further converted into object-level position annotations and pairwise direction annotations for supervising local latent learning. 
For an object with bounding box \(b_i=[x_i^{(1)}, y_i^{(1)}, x_i^{(2)}, y_i^{(2)}]\), we define its 2D reference point as the box center, i.e., \(u_i=(x_i^{(1)}+x_i^{(2)})/2\) and \(v_i=(y_i^{(1)}+y_i^{(2)})/2\). Let \(\Omega_i\) denote the set of pixels covered by the box. The object depth is estimated by averaging the predicted depth values inside the box:
\begin{equation}
z_i = \frac{1}{|\Omega_i|}\sum_{(u,v)\in\Omega_i} D(u,v).
\end{equation}

Given the camera intrinsic matrix \(K\) and pose \((R,\mathbf{t})\), where \(R\in\mathbb{R}^{3\times3}\) and \(\mathbf{t}\in\mathbb{R}^3\), the object position in the camera and world coordinates is computed as: 
\begin{equation}
\mathbf{p}_i^{c}=z_iK^{-1}
\begin{bmatrix}
u_i,
v_i,
1
\end{bmatrix}^T,
\qquad
\mathbf{p}_i^{w}=R\mathbf{p}_i^{c}+\mathbf{t}.
\end{equation}
We use \(\mathbf{p}_i^{w}\) as the object-level position annotation. For an object pair \((i,j)\), the relative displacement \(\Delta_{ij}=\mathbf{p}_j^{w}-\mathbf{p}_i^{w}\) is used to derive pairwise direction annotations.

Table~\ref{tab:pseudo_depth_3d} shows that 3D coordinates recovered from pseudo depth closely match those from ground-truth depth. The mean and median Euclidean errors are only 0.09,m and 0.04,m, with most deviation concentrated on the depth axis. Meanwhile, 98.0\% and 92.9\% of recovered points fall within 0.5,m and 0.2,m, respectively. This confirms that pseudo depth introduces only limited noise and serves as reliable 3D supervision for subsequent local latent learning.

\subsubsection{Dataset Statistics}
To increase the diversity of supervision, we construct training queries in three forms: point-coordinate queries, bounding-box-based queries, and natural-language queries. Each sample involves one to three objects depending on the task type. After filtering invalid detections and geometrically unreliable samples, the final pipeline yields 550k 3D grounding samples. Table~\ref{tab:grounding_dataset} summarizes the statistics for grounding dataset. 

\begin{table}[t]
\centering
\caption{Statistics for Spatial Reasoning datasets from SPAR.}
\label{tab:spar_dataset}
\small
\begin{tabular}{llll}
\toprule
\textbf{Source Dataset} & \textbf{Type} & \textbf{Question Type} & \textbf{Count} \\
\midrule
\multirow{8}{*}{ScanNet}
& depth\_prediction\_oc       & numeric          & 4000 \\
& depth\_prediction\_oo       & numeric          & 4000 \\
& distance\_infer\_center\_oo & multiple choice & 8000 \\
& distance\_prediction\_oc    & numeric          & 4000 \\
& distance\_prediction\_oo    & numeric          & 4000 \\
& obj\_spatial\_relation\_oo  & multiple choice & 8000 \\
& spatial\_imagination\_oc    & multiple choice & 4000 \\
& spatial\_imagination\_oo    & multiple choice & 4000 \\
\midrule
\multirow{8}{*}{ScanNet++}
& depth\_prediction\_oc       & numeric          & 4000 \\
& depth\_prediction\_oo       & numeric          & 4000 \\
& distance\_infer\_center\_oo & multiple choice & 4000 \\
& distance\_prediction\_oc    & numeric          & 4000 \\
& distance\_prediction\_oo    & numeric          & 4000 \\
& obj\_spatial\_relation\_oo  & multiple choice & 4000 \\
& spatial\_imagination\_oc    & multiple choice & 4000 \\
& spatial\_imagination\_oo    & multiple choice & 4000 \\
\midrule
\multirow{6}{*}{Structured3D}
& depth\_prediction\_oc       & numeric          & 4000 \\
& depth\_prediction\_oo       & numeric          & 4000 \\
& distance\_prediction\_oc    & numeric          & 4000 \\
& distance\_prediction\_oo    & numeric          & 8000 \\
& spatial\_imagination\_oc    & numeric          & 4000 \\
& spatial\_imagination\_oo    & numeric          & 4000 \\
\midrule
\textbf{Total} &  &  & \textbf{100000} \\
\bottomrule
\end{tabular}
\end{table}

\begin{table}[t]
\centering
\caption{Statistics for Spatial Reasoning datasets from SpatialLadder-26K.}
\label{tab:spatialladder_dataset}
\small
\begin{tabular}{lcc}
\toprule
\textbf{Question Category} & \textbf{Question Type} & \textbf{Count} \\
\midrule
Relative Direction & multiple choice & 2253 \\
Absolute Distance  & numeric         & 1127 \\
Object Size        & numeric         & 1514 \\
Relative Distance  & multiple choice & 1034 \\
\midrule
\textbf{Total}     &                 & \textbf{5928} \\
\bottomrule
\end{tabular}
\end{table}

\subsection{Spatial Reasoning Training Dataset}
The spatial reasoning training dataset is mainly built from SPAR and SpatialLadder-26K. We sample 100k questions from SPAR and 5k from SpatialLadder-26K. Basic dataset statistics are summarized in Table~\ref{tab:spar_dataset} and Table~\ref{tab:spatialladder_dataset}.

\section{Experiment Details}
\subsection{Implementation Details}
The complete implementation details are summarized in Table~\ref{tab:impl_details}. The table covers the key experimental settings for our 4-stage collaborative training.

\begin{table}[t]
\centering
\caption{Implementation details of GeoAnchor.}
\label{tab:impl_details}
\begin{tabular}{cll}
\toprule
\textbf{Category} & \textbf{Item} & \textbf{Setting} \\
\midrule

\multirow{2}{*}{General}
& Base MLLM & Qwen3-VL-2B \\
& Hardware & 8 $\times$ NVIDIA A800 \\

\midrule
\multirow{14}{*}{SFT}
& Local token length & $l_{\mathrm{pos}} = l_{\mathrm{dir}} = 2$ \\
& Global token length & $l_{\mathrm{geo}} = 8$ \\
& S1 training epochs & 1 \\
& S1 batch size & 64 \\
& S1 learning rate & $1 \times 10^{-4}$ \\
& S2/3 training epochs & 1 \\
& S2/3 batch size & 32 \\
& S2/3 learning rate & $2 \times 10^{-5}$ \\
& Text loss weight & $\lambda_t = 1$ \\
& Local loss weight & $\lambda_l = 1$ \\
& Global loss weight & $\lambda_g = 0.1$ \\
& VGGT pooling levels & $L = 3$ \\
& VGGT pooling resolutions & $\{r_1, r_2, r_3\} = \{1, 2, 4\}$ \\
& Geometry balance coefficient & $\lambda_{\mathrm{bal}} = 0.05$ \\

\midrule
\multirow{7}{*}{RL}
& Rollouts per question & $N = 8$ \\
& Sampling temperature & 1.0 \\
& Pattern reward & $r_{\mathrm{pattern}} = 0.5$ \\
& KL-divergence coefficient & $\beta = 0.01$ \\
& Learning rate & $5 \times 10^{-7}$ \\
& EMA smoothing coefficient & $\kappa = 8$ \\
& EMA update rate & $\mu = 0.2$ \\
\bottomrule
\end{tabular}%
\end{table}

\subsection{Experiment Settings}
\subsubsection{Task Definition of Position, Direction, and Mixed Questions.} 
As described in Section~\ref{sec:ablation}, the tasks from the three benchmarks are regrouped into three categories: Position, Direction, and Mixed, according to the type of local spatial evidence they require. Table~\ref{tab:task_regrouping} further provides an explicit mapping from these three categories to the original task types in each benchmark.

\begin{table}[t]
\centering
\caption{Task regrouping into Position, Direction, and Mixed categories for ablation study of local token interpretability.}
\label{tab:task_regrouping}
\begin{tabular}{lll}
\toprule
\textbf{Category} & \textbf{Task} & \textbf{Benchmark} \\
\midrule

\multirow{11}{*}{Position}
& depth\_prediction\_oc & SPAR-Bench \\
& depth\_prediction\_oo & SPAR-Bench \\
& distance\_prediction\_oc & SPAR-Bench \\
& distance\_prediction\_oo & SPAR-Bench \\
& distance\_infer\_center\_oo & SPAR-Bench \\
& obj\_spatial\_relation & SPBench \\
& object\_rel\_direction & SPBench \\
& obj\_abs\_distance & SPBench \\
& object\_size\_estimation & SPBench \\
& object\_rel\_distance & SPBench \\
& Camera - Relative Direction & ViewSpatial \\
\midrule

\multirow{2}{*}{Direction}
& Camera - Object View Orientation & ViewSpatial \\
& Person - Object View Orientation & ViewSpatial \\
\midrule

\multirow{3}{*}{Mixed}
& spatial\_imagination\_oc & SPAR-Bench \\
& spatial\_imagination\_oo & SPAR-Bench \\
& Person - Relative Direction & ViewSpatial \\
\bottomrule
\end{tabular}
\end{table}

\subsubsection{Dense Alignment Baselines for Global Geometry Supervision}
This subsection describes the three dense alignment baselines presented in ablation study of VGGT alignment strategy. 
All three methods use the same projected geometry tokens as in Section~\ref{sec:latent}, and differ only in how the dense VGGT feature sequence is compressed to match the length of the global token. Following Section~\ref{sec:latent}, let the final VGGT feature map be of size $H \times W \times D$. 
After flattening the spatial dimensions in raster order, the VGGT feature sequence is denoted as: 
\begin{equation}
F = \{f_1, \ldots, f_{l_f}\} \in \mathbb{R}^{l_f \times D}, \qquad l_f = H \times W,
\end{equation}
and the global latent $z^{\mathrm{geo}}$ is projected into geometry tokens: 
\begin{equation}
G = \{g_1, \ldots, g_{l_{\mathrm{geo}}}\} \in \mathbb{R}^{l_{\mathrm{geo}} \times D},
\end{equation}
where $l_{\mathrm{geo}}$ is the number of geometry tokens.

\begin{table*}[t]
  \centering
  \caption{\textbf{Ablation study on different base models.} GeoAnchor is instantiated on Qwen2.5-VL-3B, with comparisons against the base model, vanilla SFT, and text CoT SFT. The results show that GeoAnchor consistently achieves the best performance across all three benchmarks, demonstrating that the proposed latent reasoning framework can be effectively integrated into different base models and yields substantial gains in spatial reasoning.}
  \label{tab:ablation_arch}

  \resizebox{\textwidth}{!}{%
  \begin{tabular}{l *{12}{c}}
    \toprule
    &  \multicolumn{6}{c}{\textbf{SPAR-Bench}} & \multicolumn{3}{c}{\textbf{SPBench}} & \multicolumn{3}{c}{\textbf{ViewSpatial}} \\
    \cmidrule(lr){2-7}\cmidrule(lr){8-10}\cmidrule(lr){11-13}
    \textbf{Model}
      & \textbf{Avg.} & \textbf{Dep.} & \textbf{Dis.} & \textbf{Prox.} & \textbf{Rel.} & \textbf{View}
      & \textbf{Avg.} & \textbf{Rel.} & \textbf{Abs.}
      & \textbf{Avg.} & \textbf{Cam.} & \textbf{Per.} \\
    \midrule

    Qwen2.5-VL-3B
      & 28.7 & 25.5 & 25.7 & 55.6 & 29.4 & 21.7
      & 32.9 & 42.8 & 26.5
      & 37.3 & 39.7 & 33.6 \\
    \quad + vanilla SFT
      & 59.8 & 42.5 & 56.8 & 76.8 & 75.2 & 65.1
      & 59.2 & 68.0 & 53.5
      & 41.9 & 36.1 & 46.8 \\
    \quad + text CoT SFT
      & 60.7 & 45.2 & 60.6 & 74.2 & 76.1 & 62.5
      & 52.5 & 56.2 & 50.1
      & 39.3 & 36.8 & 43.1 \\
    \midrule
    GeoAnchor
      & \textbf{66.0} & \textbf{49.7} & \textbf{62.9} & \textbf{81.5} & \textbf{81.0} & \textbf{71.4}
      & \textbf{66.1} & \textbf{76.6} & \textbf{59.3}
      & \textbf{47.0} & \textbf{46.8} & \textbf{47.4} \\
    \textcolor{impgreen}{\textit{Improvement}}
      & \textcolor{impgreen}{+37.3} & \textcolor{impgreen}{+24.2} & \textcolor{impgreen}{+37.2} & \textcolor{impgreen}{+25.9} & \textcolor{impgreen}{+51.6} & \textcolor{impgreen}{+49.7}
      & \textcolor{impgreen}{+33.2} & \textcolor{impgreen}{+33.8} & \textcolor{impgreen}{+32.8}
      & \textcolor{impgreen}{+9.7} & \textcolor{impgreen}{+7.1} & \textcolor{impgreen}{+13.8} \\
    
    \bottomrule
  \end{tabular}%
  }
\end{table*}

\noindent\textbf{Mean Pooling.}
This baseline performs the strongest compression. 
It averages the dense VGGT sequence and the geometry-token sequence along the sequence dimension, producing one global vector for each side, and then directly aligns the two pooled vectors with Smooth L1 loss. 
In other words, the entire feature sequence is reduced to a single scene-level representation before alignment.

\noindent\textbf{Adaptive Pooling.}
This baseline first resamples the VGGT sequence from length $l_f$ to length $l_{\mathrm{geo}}$ by adaptive average pooling. 
Concretely, the sequence is partitioned into $l_{\mathrm{geo}}$ contiguous bins of nearly equal size, and the features inside each bin are averaged to obtain a resampled VGGT target sequence. 
This produces one target vector for each geometry token, enabling token-wise dense supervision after length matching.

\noindent\textbf{Linear Interpolation.}
This baseline also converts the VGGT sequence to length $l_{\mathrm{geo}}$, but uses 2D linear interpolation instead of average pooling. 
Compared with adaptive pooling, it preserves the sequence order through continuous interpolation between neighboring VGGT features, and then aligns the interpolated sequence with the geometry tokens in a token-wise manner.

For the latter two baselines, let $\tilde{F}=\{\tilde{f}_1,\ldots,\tilde{f}_{l_{\mathrm{geo}}}\}\in\mathbb{R}^{l_{\mathrm{geo}}\times D}$ denote the resampled VGGT sequence obtained by either adaptive pooling or linear interpolation. 
Their alignment loss is defined uniformly as: 
\begin{equation}
\mathcal{L}_{\mathrm{dense}}=
\frac{1}{l_{\mathrm{geo}}}
\sum_{j=1}^{l_{\mathrm{geo}}}
\mathrm{SmoothL1}(g_j,\tilde{f}_j).
\end{equation}

\section{Additional Experiments}

\subsection{Ablation Study for Different Base Model}
\definecolor{impgreen}{RGB}{255, 0, 0}

We implement our latent reasoning framework on Qwen2.5-VL-3B, and Table~\ref{tab:ablation_arch} shows the corresponding results. GeoAnchor consistently outperforms conventional SFT methods such as vanilla SFT and text CoT SFT, exhibiting the same improvement trend as on the Qwen3-VL-2B illustratede in the main paper. These results demonstrate the effectiveness and generalizability of our method, as it can be seamlessly integrated into different model backbones and consistently improve their spatial reasoning capabilities.

\begin{table*}[t]
    \centering

    \begin{minipage}[t]{0.49\textwidth}
        \vspace{0pt}
        \centering
        \captionof{table}{Ablation study on the latent length. We demonstrate that the overall model achieves the best performance when the local token length is 2 and the global token length is 8.}
        \label{tab:ablation_latent_length}

        \resizebox{\linewidth}{!}{%
            \begin{tabular}{cc|cccc}
                \toprule
                \textbf{Local} & \textbf{Global} & \textbf{SPAR} & \textbf{SPBench} & \textbf{ViewSpatial} & \textbf{Avg.} \\
                \midrule
                1 & 8 & 64.5 & 67.3 & 38.7 & 56.8 \\
                3 & 8 & 65.5 & 64.1 & 35.2 & 54.9 \\
                \midrule
                2 & 4 & 66.1 & 66.2 & 41.8 & 58.0 \\
                2 & 6 & 66.7 & 67.7 & 45.6 & 60.0 \\
                2 & 10 & 65.3 & 66.5 & 43.2 & 58.3 \\
                2 & 12 & 65.7 & 66.4 & 41.6 & 57.9 \\
                \midrule
                2 & 8 & \textbf{67.5} & \textbf{68.8} & \textbf{46.3} & \textbf{60.9} \\
                \bottomrule
            \end{tabular}%
        }
    \end{minipage}
    \hfill
    \begin{minipage}[t]{0.49\textwidth}
        \vspace{0pt}
        \centering
        \captionof{table}{Ablation study of pooling resolution for VGGT alignment. The multi-scale setting \(\{1,2,4\}\) performs best among all multi-scale variants, and further surpasses the single-scale \(5\times5\) setting even with a comparable token count.}
        \label{tab:ablation_sampleing}

        \resizebox{\linewidth}{!}{%
            \begin{tabular}{c|cccc}
                \toprule
                \textbf{Pooling Resolution} & \textbf{SPAR} & \textbf{SPBench} & \textbf{ViewSpatial} & \textbf{Avg.} \\
                \midrule
                \{1, 2\} & 65.2 & 67.1 & 41.6 & 57.9 \\
                \{1, 2, 4, 8\} & 64.3 & 66.8 & 44.3 & 58.5 \\
                \{5\} & 67.2 & 66.9 & 45.9 & 60.0 \\
                \midrule
                \{1, 2, 4\} & \textbf{67.5} & \textbf{68.8} & \textbf{46.3} & \textbf{60.9} \\
                \bottomrule
            \end{tabular}%
        }
    \end{minipage}

\end{table*}

\subsection{Ablation Study for Different Latent Length}
We carefully tune the lengths of the local and global tokens to ensure that the latent space remains both expressive and compact. Table~\ref{tab:ablation_latent_length} shows that the best overall performance is achieved when the local token length is 2 and the global token length is 8. This indicates that effective spatial reasoning depends on a latent space that remains compact while preserving sufficient spatial information. Overly short latents impose an excessive information bottleneck, while overly long latents weaken the compactness and functional specialization of the latent space. The balanced setting therefore provides the most effective trade-off between preserving spatial information and supporting downstream reasoning.

\subsection{Ablation Study for Different Pooling Resolution in Global Token Alignment}

Table~\ref{tab:ablation_sampleing} shows that the multi-scale setting \(\{1,2,4\}\) achieves the best overall performance, obtaining the best results on all three benchmarks. This indicates that combining coarse-to-fine spatial resolutions provides more effective geometry supervision than either a smaller or an overly fine scale set. Moreover, although the single-scale \(5\times5\) setting uses a comparable number of tokens, it still underperforms \(\{1,2,4\}\). This suggests that the advantage of multi-scale pooling lies in its ability to capture complementary geometric cues at different spatial resolutions, jointly preserving global scene layout and finer local structure, which leads to more effective geometry alignment.

\end{document}